\documentclass[10pt,twocolumn,letterpaper]{article}

\usepackage{cvpr}              

\usepackage{graphicx}
\usepackage{amsmath}
\usepackage{amssymb}
\usepackage{booktabs}
\usepackage{url}            
\usepackage{booktabs}       
\usepackage{amsfonts}       
\usepackage{nicefrac}       
\usepackage{microtype}      
\usepackage{graphicx}
\usepackage{multirow}
\usepackage{wrapfig}
\usepackage{xcolor}
\usepackage{bbm}
\usepackage{amsmath}
\usepackage{comment}
\usepackage{bbm}
\usepackage{footnote}
\usepackage{algpseudocode}
\usepackage[accsupp]{axessibility}
\usepackage[linesnumbered,boxed,ruled,commentsnumbered]{algorithm2e}
\usepackage{footnote}
\newcommand\blfootnote[1]{%
  \begingroup
  \renewcommand\thefootnote{}\footnote{#1}%
  \addtocounter{footnote}{-1}%
  \endgroup
}

\usepackage[pagebackref,breaklinks,colorlinks]{hyperref}

\usepackage[capitalize]{cleveref}
\crefname{section}{Sec.}{Secs.}
\Crefname{section}{Section}{Sections}
\Crefname{table}{Table}{Tables}
\crefname{table}{Tab.}{Tabs.}

\newcommand{\phil}[1]{{\color{orange}{#1}}}

\def\cvprPaperID{2043} 
\def\confName{CVPR}
\def\confYear{2022}

\begin{document}

\title{Towards Implicit Text-Guided 3D Shape Generation}

\author{Zhengzhe Liu$^{1}$ \quad   Yi Wang$^{2}$ \quad    Xiaojuan Qi$^{3*}$   \quad   Chi-Wing Fu$^{1*}$ \\
$^1$The Chinese University of Hong Kong \quad $^2$Shanghai AI Laboratory \quad $^3$The University of Hong Kong \\
{\tt\small \{zzliu,cwfu\}@cse.cuhk.edu.hk \quad wangyi@pjlab.org.cn \quad  xjqi@eee.hku.hk }
}
\maketitle

\begin{abstract}
\blfootnote{*: Corresponding authors}
In this work, we explore the challenging task of generating 3D shapes from text.
Beyond the existing works, 
we propose a new approach for text-guided 3D shape generation, capable of producing high-fidelity shapes with colors that match the given text description.
This work has several technical contributions.
%
First, we decouple the shape and color predictions for learning features in both texts and shapes, and 
%
propose the word-level spatial transformer to correlate word features from text with spatial features from shape.
Also, we design a cyclic loss to encourage consistency between text and shape, and introduce the shape IMLE to diversify the generated shapes.
Further, we extend the framework to enable text-guided shape manipulation. 
Extensive experiments on the largest existing text-shape benchmark~\cite{chen2018text2shape} manifest the superiority of this work. The code and the models are available at {\footnotesize \url{https://github.com/liuzhengzhe/Towards-Implicit-Text-Guided-Shape-Generation}}.
\end{abstract}

\vspace{-5mm}
\section{Introduction}
\label{sec:intro}

3D shape creation has a wide range of applications,~\eg, CAD, games, animations, computational design, augmented reality, etc.
Significant progress has been made in recent years by exploiting neural networks and generative models to learn to produce 3D shapes.
Yet, existing works~\cite{mo2019structurenet,gao2019sdm,xu2019disn,chen2019learning,chen2020bsp,jiang2020shapeflow,cai2020learning,tretschk2020patchnets,jiang2020local,liu2021deep,Jie20DsgNet} focus mostly on generating the overall shapes, whereas the more recent ones~\cite{li2020d,genova2020local,yariv2020multiview,poursaeed2020coupling,yifan2021iso,chen2021decor,chen2021multiresolution} attempt to generate shapes with more details.

\begin{figure}
\centering
\includegraphics[width=0.98\columnwidth]{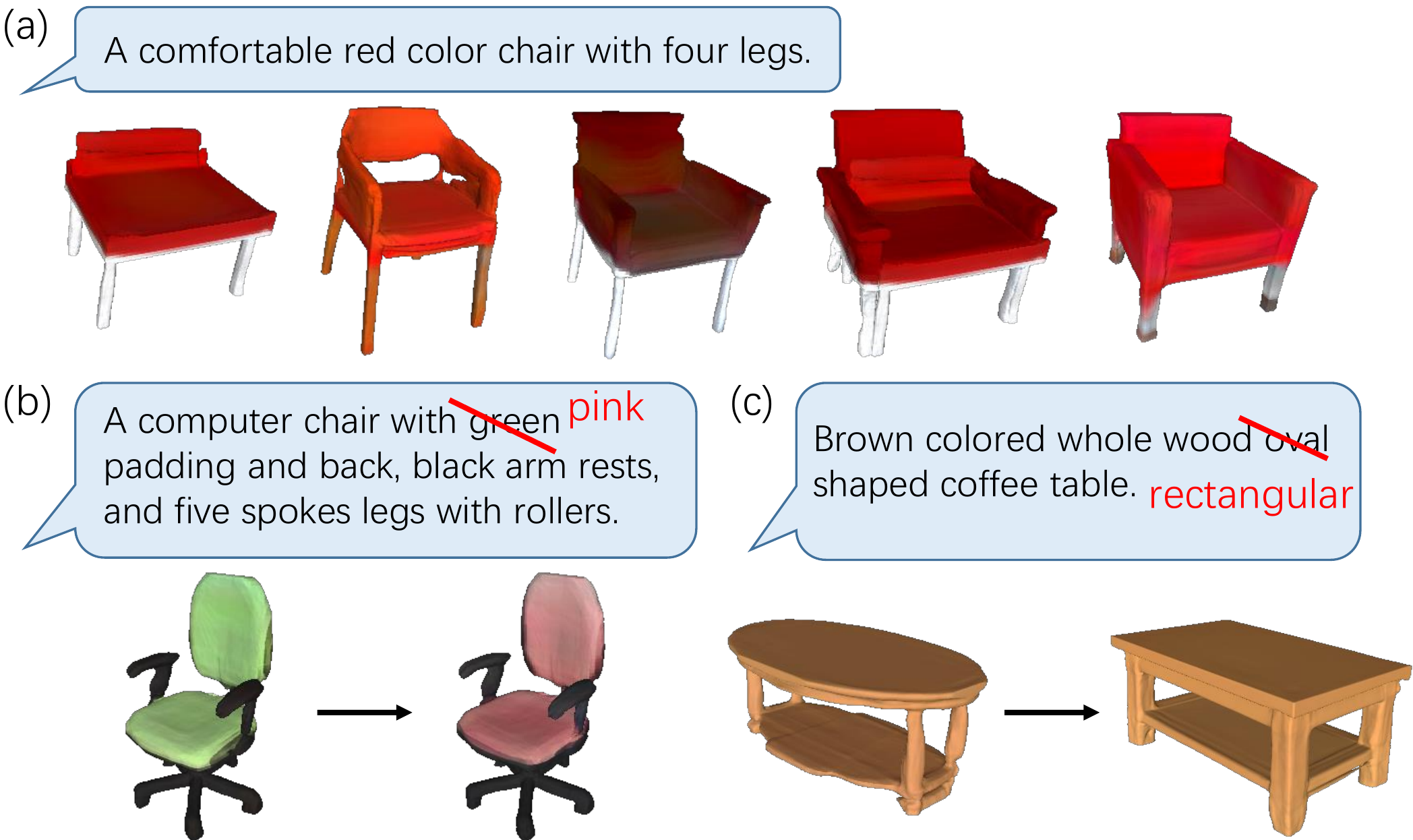}
\vspace*{-2mm}
\caption{(a) Chairs of different structures and appearances generated by our method from the same given sentence.
Our method also allows text-based manipulation in color (b) and in shape (c).
%
}
\label{fig:demo}
\vspace*{-3mm}
\end{figure}

In this work, we are interested in the challenging task of {\em text-guided 3D shape generation\/}---Given a sentence, e.g., ``A comfortable red color
chair with four legs,'' we aim to develop a method to automatically generate a 3D shape that follows the text description; see Figure~\ref{fig:demo} (a) for our example results.
This research direction has great potential for efficient 3D shape production, say by taking user speech/text input to guide or condition the process of generating 3D shapes.
By this means, we can assist users to readily generate and edit 3D models for diverse applications.


While many methods~\cite{reed2016generative,reed2016learning,zhang2017stackgan,zhang2018stackgan++,xu2018attngan,li2019controllable,qiao2019mirrorgan,stap2020conditional,wang2020text,rombach2020network,wang2021cycle} have been developed for generating 2D images from text, the task of generating 3D shapes from text is rather under-explored. 
Chen~\etal~\cite{chen2018text2shape} generate 3D shapes from natural language descriptions by learning joint text and shape embeddings, but the performance and visual quality are highly limited by the low-resolution 3D representations.
Another very recent work~\cite{jahan2021semantics} leverages semantic labels to guide the shape generation, but it requires predefined semantic labels and cannot directly deal with natural language inputs.

To enhance 3D shape generation from text, we propose a new solution by  
leveraging the implicit representation~\cite{mescheder2019occupancy,chen2019learning,park2019deepsdf} to predict an occupancy field.
Yet, several inherited challenges have not been addressed in the early works for properly adopting the implicit representation for the text-to-shape task.
First, the above works generate shapes typically without colors, which are crucial in text-guided 3D shape generation, since text descriptions often contain colors; we empirically found that directly predicting shape and color with a single implicit decoder often lead to shape distortion and color blur. 
Second, text contains a large amount of spatial-relation information,~\eg,~``a wooden table on a metal base.''
Still, spatial-relation local features are ignored in existing works, since the implicit decoder generally considers only the global feature from the auto encoder as input~\cite{chen2019learning}.
%
%
Third, the generated shapes are not all consistent with the input texts, largely due to
the semantic gap between text and 3D shape and also the lack of effective learning constraints.
Last, text-to-shape generation is inherently one-to-many,~\ie, diverse results may match the same input text.
Yet,  the existing regression-based approach outputs only a single shape.




This work presents a new approach for high-fidelity text-guided 3D shape generation.
First, we decouple the shape and color predictions for feature learning in both texts and shapes to improve the generation fidelity; this strategy also aids the text-guided shape manipulation.
Also, we introduce a word-level spatial transformer to learn to correlate the word features with the spatial domain in shapes.
In addition, we design a cyclic loss to encourage the consistency between the generated 3D shape and the input text. 
Further, we propose a novel style-based latent shape-IMLE generator for producing diversified shapes from the same given text. 
%
%
%
Last, we extend the framework for text-guided 3D shape manipulation with a two-way cyclic loss.
As shown in Figure~\ref{fig:demo} (b), we may modify the original text and our framework can produce new colored shapes according to the edited text, while keeping the other attributes unchanged.

Extensive experiments on the largest existing text-shape dataset~\cite{chen2018text2shape} demonstrate the superiority of our approach over the existing works, both qualitatively and quantitatively.



\begin{figure*}
\centering
\includegraphics[width=0.99\textwidth]{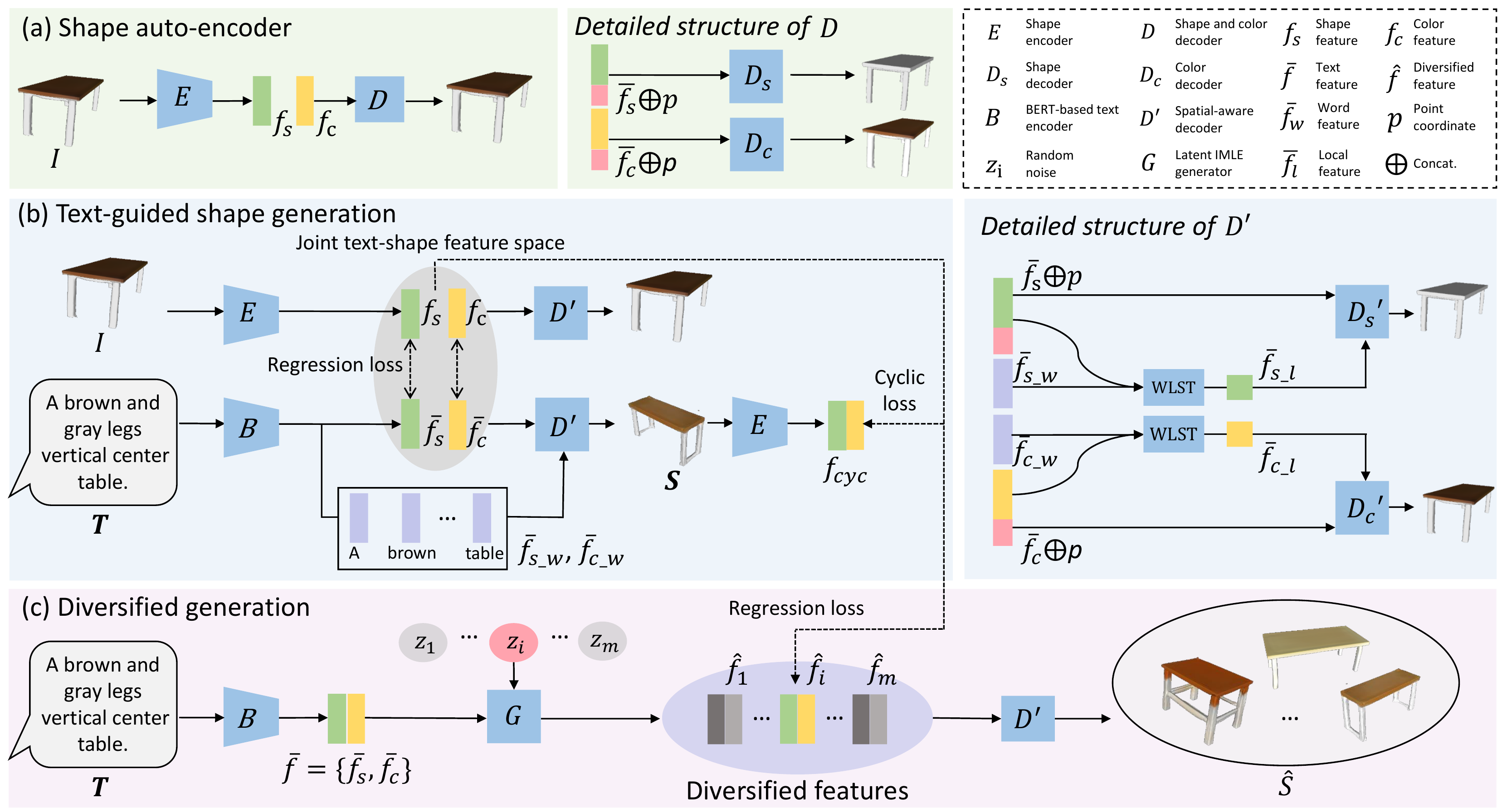}
\vspace*{-1.5mm}
\caption{Overview of our text-guided shape generation framework, which has three major parts.
(a) First, the shape auto-encoder $\{E,D\}$ extracts shape feature $f_s$ and color feature $f_c$ from the input 3D shape $I$.
(b) We then learn to generate the 3D shape in a text-guided manner with the word-level spatial transformer (WLST) and the cyclic consistency loss $f_{cyc}$.
(c) Further, we generate diversified 3D shapes from the same given text by adopting a style-based latent shape generator $G$.
We only need (c) during the inference.
}
\label{fig:overview}
\vspace*{-2.5mm}
\end{figure*}

\vspace{-1.5mm}
\section{Related Work}
\label{sec:RW}

\paragraph{Text-to-image generation.}
Remarkable progress has been made for generating images from text~\cite{reed2016generative,reed2016learning,zhang2017stackgan,zhang2018stackgan++,xu2018attngan,li2019controllable,li2020manigan,qiao2019mirrorgan,wang2021cycle}.
%
%
Recently, approaches~\cite{stap2020conditional,
yuan2019bridge,
souza2020efficient,
wang2020text,
rombach2020network,
patashnik2021styleclip,
xia2021tedigan} based on the unconditional GAN~\cite{karras2019style,karras2017progressive,brock2018large} were also proposed.

Compared with text-to-image, it is more challenging to generate 3D shapes from texts. First, unlike 2D images, 3D shapes are unstructured and irregular without well-defined grid structures. Also, the text-to-shape task requires a comprehensive prediction of the whole 3D shape, while the text-to-image task addresses image generation, which is a projection of the 3D shape.
Further, there are plenty of large-scale image datasets~\cite{wah2011caltech,nilsback2008automated,lin2014microsoft} to support text-to-image.
Yet, as far as we know, the largest dataset for text-to-shape was proposed in~\cite{chen2018text2shape}, which has $75k$ texts and $15k$ shapes of $128^3$ resolution.
The lack of large-scale and high-quality training data makes the text-to-shape task even harder.



\vspace*{-4mm}
\paragraph{3D shape representations, generation, and manipulation.}
Unlike images, 3D shapes can be represented as,~\eg, voxel grids~\cite{choy20163d,girdhar2016learning}, point
clouds~\cite{achlioptas2018learning2,qi2017pointnet}, and meshes~\cite{feng2019meshnet}.
Also, various methods~\cite{sun2020pointgrow,hui2020progressive,kim2020softflow,klokov2020discrete,li2021sp} have been proposed for generating and manipulating shapes for different 3D representations.
Yet, the generated shapes are limited by the resolution and quality of the training set.
To generate shapes of arbitrary resolution, recent works~\cite{mescheder2019occupancy,chen2019learning,park2019deepsdf,liu2021deep,chibane2020neural} start to explore implicit functions, which in fact have been used in many tasks,~\eg, single-view reconstruction~\cite{xu2019disn,li2020d,niemeyer2020differentiable}, 3D scene reconstruction~\cite{jiang2020local,peng2020convolutional,huang2021di}, and 3D texture generation~\cite{oechsle2019texture,chibane2020implicit2,oechsle2020learning}. In existing works, a typical approach is to leverage an auto-encoder (AE) to adopt to multiple 3D generation tasks and map the input modalities into the AE's learned feature space,~\eg, single-image 3D reconstruction~\cite{chen2019learning,xu2019disn}, point-cloud-based shape generation~\cite{cai2020learning,chibane2020implicit}, and 3D completion \cite{wu2020pq}.

Following the above works, a straightforward approach for text-guided 3D shape generation is to map the text feature into the AE's feature space then adopt an implicit decoder to generate the 3D shape. This simple approach, however, has several drawbacks, as discussed in Section~\ref{sec:intro}.

Recently, several works make it possible to manipulate implicit 3D shapes~\cite{hao2020dualsdf,ibing20213d,deng2021deformed,zheng2021deep} using a reference box or reference points as guidance.
Yet, none of them enables 3D shape manipulation with natural language descriptions.


\vspace*{-4mm}
\paragraph{3D shape generation from text.}
A series of works are proposed to address the tasks on texts and 3D shapes, including learning the text-shape correspondence~\cite{achlioptas2018learning}, cross-modal retrieval~\cite{han2019y2seq2seq,tang2021part2word}, shape-to-text generation~\cite{han2020shapecaptioner}, text-guided shape composition~\cite{huq2020static}, and 3D object localization~\cite{chen2020scanrefer}. 

As far as we are aware of, there are only few works~\cite{jahan2021semantics,chen2018text2shape} that address the challenging text-to-shape task.
%
Chen~\etal~\cite{chen2018text2shape} propose to directly predict colored voxels with adversarial learning on top of a jointly-learned text-shape embedding. 
Though plausible shapes can be produced, the shape resolution and texture quality are still far from being satisfactory. 
Also, the generated shapes may not be consistent with the input texts due to the large semantic gap between text and shape.
Jahan~\etal~\cite{jahan2021semantics} propose a semantic-label guided shape generation approach; however, it can only take one-hot semantic keywords as input and the generated shapes are also unsatisfying in quality, without color and texture.

This work presents a new framework, capable of generating high-fidelity 3D shapes with good semantic correspondence between the text and shape.
Also, our framework enables text-guided 3D shape manipulation for both shape and color, outperforming the existing works by a large margin, as demonstrated in the experiments.

\vspace*{-4mm}
\paragraph{Diversified generation.}
%
Besides GANs, IMLE (Implicit Maximum Likelihood Estimation) is another approach to aid multi-modal generation,~\eg, super-resolution~\cite{li2020multimodal}, semantic-layout-guided image synthesis~\cite{li2019diverse}, image decompression~\cite{peng2020generating}, and shape completion~\cite{arora2021shape}. Compared with GANs, IMLE mitigates the mode collapse of GANs and boosts the result diversity.
In this work, we leverage IMLE for generating multiple shapes from the same text input.


\section{Methodology}

\vspace*{-1mm}
\subsection{Overview}
Given text $\mathbf{T}$, we aim to generate high-quality 3D shape $\mathbf{S}$ with colors, following the description of $\mathbf{T}$.
To generate high-quality results, we exploit the implicit occupancy representation other than the explicit voxel/point/mesh representations to characterize shapes with color. Specifically, the predicted shape with color is denoted as $\mathbf{S} \in \mathbb{R}^{N \times (1+3)}$, including the shape $\in \mathbb{R}^{N \times 1}$(a set of occupancy values in the voxels) and the color $\in \mathbb{R}^{3N}$ (the associated set of RGB values), respectively, where $N$ is the number of sample points, concerning the generation quality.

Our framework consists of a text encoder $B$, feature generator $G$, spatial aware decoder $D'$, and shape encoder $E$. Its overall architecture is given in Figure~\ref{fig:overview}. 
In inference, $B$ extracts text feature $\bar{f}=\{\bar{f}_s,\bar{f}_c\}$ from text $\mathbf{T}$ (where $\bar{f}_s$ and $\bar{f}_c$ are the shape and color portions of $\bar{f}$, respectively), and $G$ produces multiple instances of such feature $\{\hat{f}_i\}^m_{i=1}$ based on $\bar{f}$ conditioned on various random vectors $\{z_i\}$. Then, $D'$ generates diverse shapes $\{\mathbf{S}_i \in \mathbb{R}^{N \times (1+3)}\}^m_{i=1}$ with color.

The model training of our method is non-trivial. We train the overall framework in {\em three\/} stages (see again Figure~\ref{fig:overview}): (a) shape auto-encoder, (b) text-guided shape generation, and (c) diversified shape generation with IMLE. Specifically,
\begin{itemize}
\vspace*{-2mm}
\item
First, as shown in Figure~\ref{fig:overview} (a), we train shape encoder $E$ and implicit decoder $D$.
As shown in the top middle, unlike existing works~\cite{chen2019learning,jahan2021semantics} that ignore colors in the shape generation, $D$ composes of $D_s$ and $D_c$ that account for the decoding of shape and color, respectively, when $D$ predicts the output shape.
\vspace*{-2mm}
\item
Then, we adopt BERT-based text encoder $B$~\cite{devlin2018bert} to help extract text feature $\bar{f} = \{\bar{f}_{s}, \bar{f}_{c}\}$ and word-level feature $\bar{f}_{w} = \{\bar{f}_{s\_w}, \bar{f}_{c\_w}\}$ from input text $\mathbf{T}$ (see Figure~\ref{fig:overview} (b)), and map $\bar{f}$ into the joint text-shape feature space to reduce the domain gap between the text and the shape.
Further, we propose the spatial-aware decoder $D^\prime$ to leverage local feature $\bar{f}_l$ extracted by the word-level spatial transformer (WLST), which explicitly correlates the spatial and word features to improve the fidelity of $\textbf{S}$.
Also, we formulate cyclic loss $L_{cyc}$ to encourage the consistency between shape $\mathbf{S}$ and text $\mathbf{T}$.
\vspace*{-2mm}
\item
Lastly, we propose to adopt style-based shape generator $G$ that conditions on a set of random noise vectors 
$\{ z_i \}^m_{i=1}$
to enable diversified 3D shape generation with feature $\hat{f}_i$, as shown in Figure~\ref{fig:overview} (c).
\end{itemize}
%


In the following, we will detail each component of the framework and the associated losses.



\subsection{Shape Auto-Encoder}~\label{sec:ae}
We extend the auto-encoder in~\cite{chen2019learning} 
to jointly reconstruct the shape and color.
As shown in Figure~\ref{fig:overview} (a), our shape auto-encoder aims to map the input voxel-based shape $I \in \mathbb{R}^{64 \times 64 \times 64}$ into a compact feature space.
Specifically, encoder $E$~\cite{chen2019learning} extracts the shape and color features $f=\{f_s,f_c\}$ from $I$, 
whereas decoder $D$ reconstructs the shape and color through $D_s$ and $D_c$, respectively.
Inside $D$, we concatenate a sample (or query) point coordinate $p=(x,y,z)$ with each feature vector ($f_s$ or $f_c$) as input 
to $D_s$ or $D_c$.
$D_s$ and $D_c$ have the same architecture with seven fully-connected and leaky-ReLU layers, except in the last layer, $D_s$ outputs a single occupancy value and $D_c$ outputs three values for RGB color, both at the sample point $p$.

The shape auto-encoder is trained to reconstruct the shape and color of the input shape with an $L_2$ regression:
\vspace*{-1mm}
\begin{equation}
\begin{aligned}
L_{ae}=&\lambda_s \Sigma_{p}||D_s(f_s \oplus p)-I(p)||^2_2\\
+&\lambda_c \Sigma_{k\in\{R,G,B\}}\Sigma_{p}||D_c(f_c \oplus p)[k] - I(p)[k]||^2_2\mathbbm{1}(I(p)),
\label{equ:ae}
\end{aligned}
\end{equation}
\noindent
where $I(p)$ and $I(k,p)$ denote the ground-truth occupancy and color values, respectively, at point $p$;
$\oplus$ denotes concatenation;
$\mathbbm{1}$ is an indicator function of value $1$ if $p$ is inside the input shape, and $0$, otherwise; and
$\lambda_s$ and $\lambda_c$ are weights for the shape and color reconstructions, respectively.




\subsection{Text-Guided Shape Generation}

As shown in Figure~\ref{fig:overview} (b), the text-guided shape generation network consists of three modules: shape encoder $E$, BERT-based text encoder $B$, and spatial-aware decoder $D^\prime$.
With $E$ and $D^\prime$ ($D_s^\prime$ and $D_c^\prime$) initialized by the corresponding components in the shape auto-encoder, our goal here is to train the whole network end-to-end to obtain $B$ and $D^\prime$.

\vspace*{-3.5mm}
\paragraph{Text encoder $B$.}
We employ the BERT structure~\cite{devlin2018bert} to build text encoder $B$ for extracting text feature $\bar{f}$ from input text $\mathbf{T}$ and mapping $\bar{f}$ to the joint text-shape feature space.


\begin{figure}[!t]
\centering
\includegraphics[width=0.99\columnwidth]{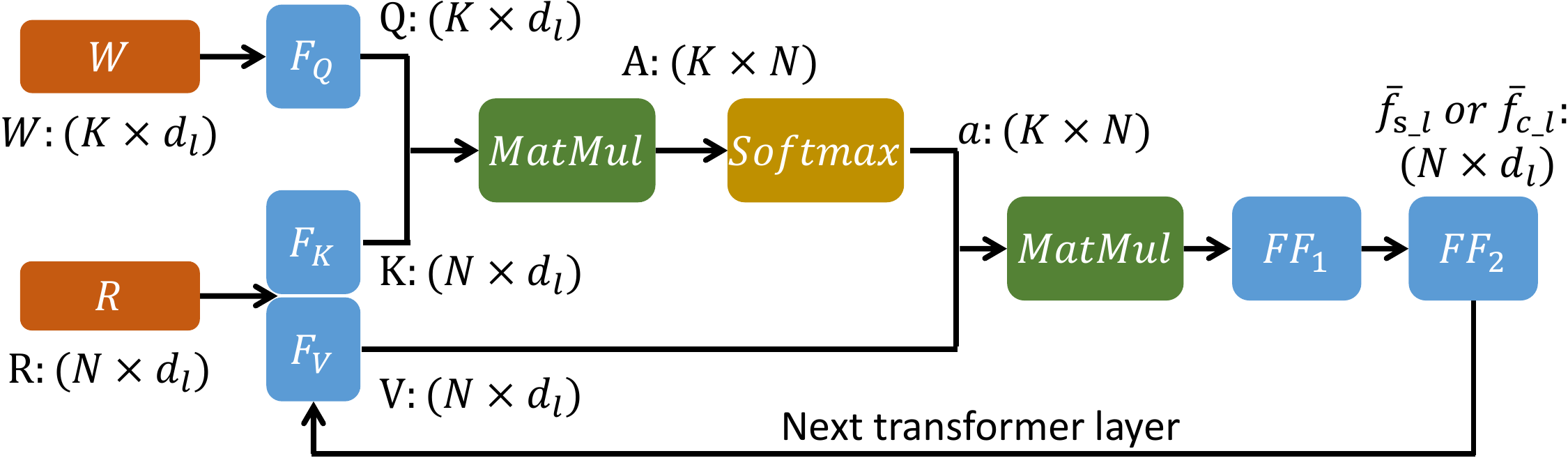}
\vspace*{-1mm}
\caption{The Word-Level Spatial Transformer architecture.
$F_Q$, $F_K$, and $F_V$ are fully-connected layers, whereas $FF_1$ and $FF_2$ are feed-forward networks.
The Layer Normalization~\cite{ba2016layer} is omitted.
}
\label{fig:transformer}
\vspace*{-2mm}
\end{figure}

\begin{figure}[!t]
\centering
\includegraphics[width=0.99\columnwidth]{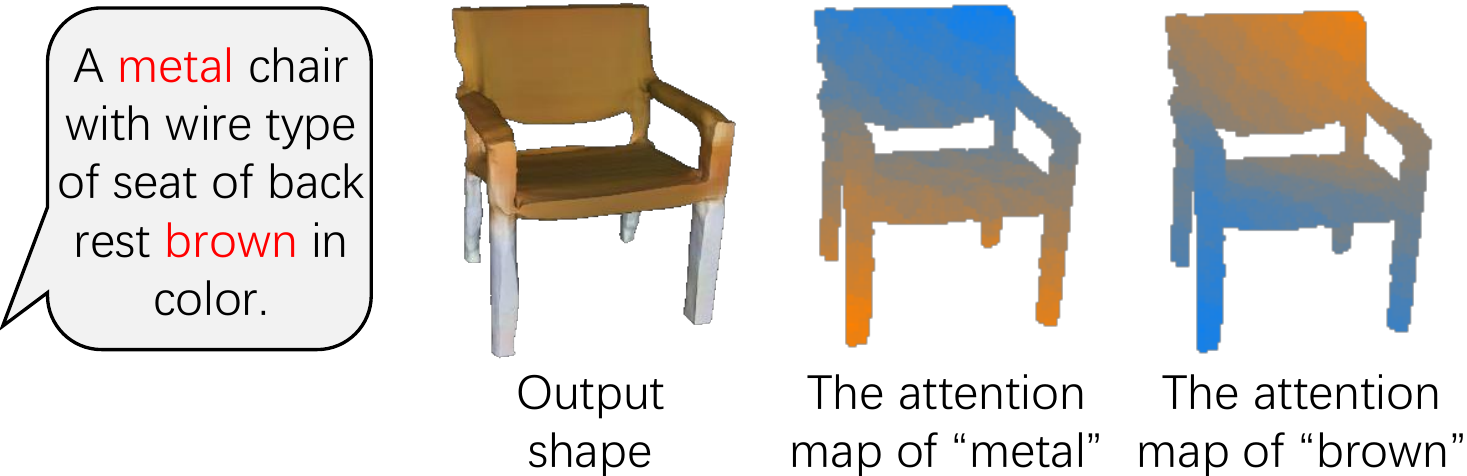}
\vspace*{-2mm}
\caption{Visualizing the attention map $A$ for the words ``metal'' and ``brown''.
Warmer colors indicate stronger correlation. 
}
\label{fig:attention}
\vspace*{-3mm}
\end{figure}

\vspace*{-3.5mm}
\paragraph{Spatial-aware decoder $D^\prime$.}
$D^\prime$ aims to transform text feature $\bar{f}$ to the predicted shape $\mathbf{S}$ with color. Instead of simply using the trained implicit decoder $D$, we construct the spatial-aware decoder $D^\prime$ with the word-level spatial transformer (WLST).
In short, we take the local features from WLST to improve the spatial correlation implied from $\mathbf{T}$.


The right side of Figure~\ref{fig:overview}(b) shows the architecture of the spatial-aware decoder $D^\prime$.
First, we concatenate $\bar{f}_s$ and $p$ and transform the result $\{ \bar{f}_s \oplus p \}$ $\in\mathbb{R}^{N \times (d+3)}$ using a fully-connected layer, where $N$ is the number of sample points for shape reconstruction and $d$ is the channel dimension of $\bar{f}_s$.
Then, we transform the word-level BERT features $\{ \bar{f}_w \} \in \mathbb{R}^{K \times d_{B}}$ (where $K$ is the number of words in input text and $d_{B}$ is the channel dimension of each word feature $\bar{f}_w$) from $B$ using a fully-connected layer.
%
%
The transformed spatial and word features are denoted as $R\in\mathbb{R}^{N \times d_l}$ and $W\in\mathbb{R}^{K \times d_l}$, respectively,
%
where $R_i \in \mathbb{R}^{d_l}$ is the $i^{th}$ row in $R$ that corresponds to the $i^{th}$ sample point and $W_j \in \mathbb{R}^{d_l}$ is the $j^{th}$ row in $W$ that corresponds to the $j^{th}$ word in input text.
Importantly, we formulate the WLST to learn the correlation between $\{R_i\}$ and $\{W_j\}$; see the next paragraph for the details.
After that, $D^\prime_s$ takes the global feature $\bar{f}_s$, sample point coordinate $p_i$, and local feature $\bar{f}_{s\_l,i}$ from WLST as inputs to predict the occupancy value at $p_i$ for shape reconstruction.

Figure~\ref{fig:transformer} shows the architecture of the WLST. With the spatial features $R$ and word features $W$, we first establish an attention map $A$ to explicitly correlate each word feature $W_j$ with each sample point $p_i$ given the shape feature $\bar{f}_s$; see Figure~\ref{fig:attention} for example visualizations of $A$, revealing how it captures the spatial regions in a shape for different words in the input text.
Next, we use the $softmax$ function to process $A$ to generate the normalized attention matrix $a$. The output local shape feature $\bar{f}_{s\_l,i}$
of point $p_i$ is the weighted aggregation of the word-level features $W_j$ across the whole input text.
%
%
Hence, our WLST can be formulated as 
\begin{equation} 
\bar{f}_{s\_l,i}=\Sigma_j softmax(\frac{F_Q(W_j)F_K(R_i)}{\sqrt{d_l}})F_V(R_i),
\label{equ:transformer}
\end{equation}
\noindent
where $F_Q$, $F_K$, and $F_V$ are fully-connected layers; see Figure~\ref{fig:transformer} for the architecture of the WLST.
Similarly, $D^\prime_c$ also leverages a WLST for extracting local color feature $\bar{f}_{c\_l}$.

With the WLST, we can extend the implicit decoder $D$ to take into account the extra local feature $\bar{f}_l = \{ \bar{f}_{s\_l}, \bar{f}_{c\_l} \}$ (see Figure~\ref{fig:overview}), which is produced by explicitly learning the correlation between the word-level spatial descriptions and the 3D shape.
Hence, we can make every single word in the input text accessible to the shape decoder and enhance the fidelity (or local details) of the generated shape.

\vspace*{-3mm}
\paragraph{Cyclic consistency loss.}
To reduce the semantic gap between the text and shape, we propose a cyclic consistency loss to encourage the consistency between input text $\textbf{T}$ and output shape $\textbf{S}$ from $D^\prime$.
%
%
To form a cycle, we first grid-sample $64\times64\times64$ points to use $D^\prime$ to generate $\textbf{S}$, and utilize encoder $E$ from the trained shape auto-encoder to extract features $f_{cyc}$ from $\textbf{S}$; see Figure~\ref{fig:overview}(b).
Then, we define the cyclic consistency loss to operate on the semantic meaningful feature space instead of the low-level occupancy or color values, such that it can regularize the shape generation in a closed loop by encouraging the high-level features $f_{cyc}$ to be similar to $f = \{f_s,f_c\}$ from the shape encoder.

To reduce the memory consumption and training time, we firstly grid-sample $16\times16\times16$ points to form a low-resolution voxelized shape $S_l$, then tri-linearly upsample $S_l$ to $\textbf{S}$ of the same resolution as $I$ ($64\times64\times64$).



\vspace*{-3mm}
\paragraph{Network training.}
Initialized with the shape auto-encoder, we train the text-guided shape generation network end-to-end with the shape auto-encoder loss $L_{ae}^\prime$ on $D^\prime$, 
%
%
\begin{equation}
\begin{aligned}
 \resizebox{0.87\columnwidth}{!}{$ \begin{split}
L_{ae}^\prime=&\lambda_s \Sigma_{p}||D_s^{\prime}(f_s,p,\bar{f}_{s\_l},R_{s\_i})-I(p)||^2_2\\
+&\lambda_c \Sigma_{k\in\{R,G,B\}}\Sigma_{p}||D_c^{\prime}(f_c,p,\bar{f}_{c\_l})[k]-I(p)[k]||^2_2\mathbbm{1}(I(p)),
\end{split}$}
\label{equ:ae}
\end{aligned}
\end{equation}
\vspace{-1mm}
%
%
\begin{equation}
L_{reg}=\lambda_r ||\bar{f}- f||^2_2,
\label{equ:reg}
\end{equation}
\begin{equation} 
\text{and} \
L_{cyc}=
\lambda_{cyc}||f_{cyc}- f||^2_2,
\label{equ:cyc}
\end{equation}
where $\lambda_s$, $\lambda_c$, 
$\lambda_r$, and $\lambda_{cyc}$ are weights.


\subsection{Diversified 3D Shape Generation}
\begin{figure}
\centering
\includegraphics[width=0.99\columnwidth]{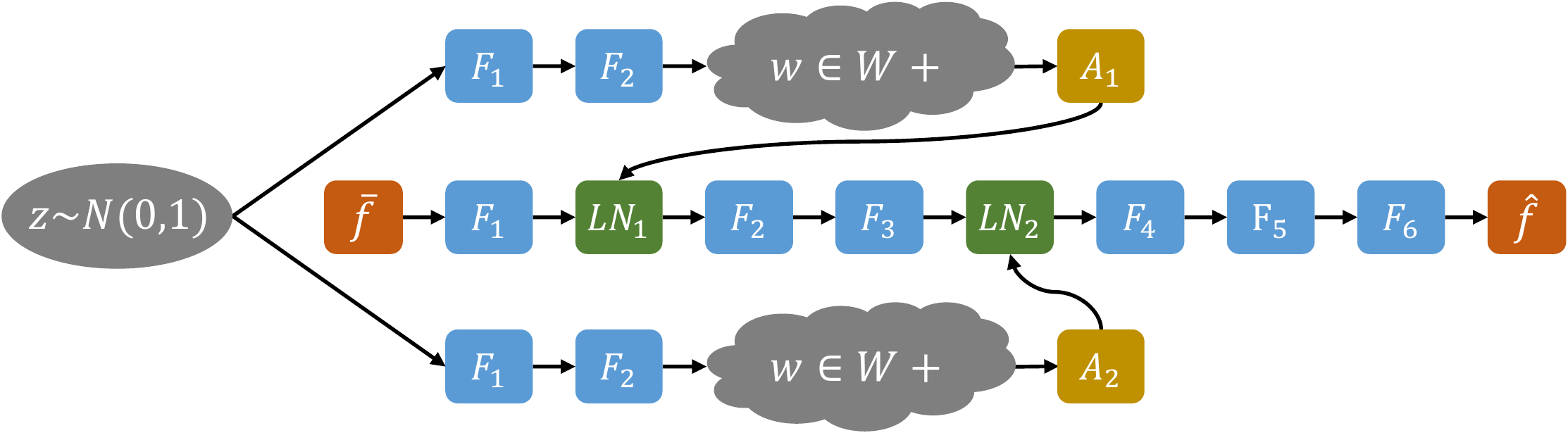}
\caption{The architecture of our shape-IMLE generator.
Inspired by StyleGAN~\cite{karras2019style}, we map random noise $z$ to latent space W+~\cite{abdal2019image2stylegan} to control the generator through adaptive Layer Normalization~\cite{ba2016layer} ($A_1$ and $A_2$) at the first and third fully-connected layers.
}
\label{fig:generator}
\vspace*{-2mm}
\end{figure}

To enable diversified 3D shape generation for the same input text, we propose a style-based latent shape-IMLE generator $G$, namely shape IMLE, which operates in the latent space; see Figure~\ref{fig:overview}(c).
Taking text feature $\bar{f} = \bar{f}_s \oplus \bar{f}_c$ from $\mathbf{B}$ as input, $G$ generates $\{\hat{f}_i = \hat{f}_{s,i} \oplus \hat{f}_{c,i} \}^m_{i=1}$ conditioned on a set of random vectors $Z=\{z_i\}_{i=1}^m$.
Different from GANs, which encourage the generated samples to be similar to the target data, IMLE inversely encourages each target data to have a similar generated sample to avoid mode collapse~\cite{li2019diverse}. 
Figure~\ref{fig:generator} shows the architecture of the shape IMLE $G$.

For the training of $G$, it is optimized as follows:
%
%
\begin{equation}
\begin{aligned}
\min\limits_{\theta}\mathbbm{E}_Z[\min\limits_{k\in\{1,\dots,m\}}d(G_\theta(\bar{f},z_{k}), f)]
\label{equ:imle}
\end{aligned}
\end{equation}
%
\noindent where $\theta$ denotes the weights of generator $G$; $d(\cdot,\cdot)$ is a distance metric; and $z_{k} \sim N(0,1)$.

With each input $\bar{f}$, we randomly sample $m$ random noise vectors $\{z_i\}$ to generate $m$ different outputs $\{\hat{f}_i\}$. Among them, the one that is most similar to the ground truth $f$, say $\hat{f}_{k}$, is trained to be closer to $f$ 
with an $L_2$ regression.
So, we can encourage every 
ground truth $f$ 
to have a similar generated sample 
to avoid the mode-collapse issue in GANs, while promoting diversified shape generation~\cite{li2019diverse,li2020multimodal}.
%
%
We train the shape IMLE $G$ with all the other modules $E,B,D^\prime$ frozen (see Figure~\ref{fig:overview})
using an $L_2$ loss on $\hat{f}_{k}=G(\bar{f},z_{k})$:
\begin{equation} 
\begin{aligned}
L_{G}= \min\limits_{k\in\{1,\dots,m\}} ||G(\bar{f},z_{k}), f ||_2^2.
\label{equ:imle}
\end{aligned}
\end{equation}

During the inference, we feed every feature of $\{\hat{f}_{1},\dots,\hat{f}_{m}\}$ into $D$ for generating diversified shapes, without using the ground truth $f$ 
to select the nearest $\hat{f}_{k}$.


\begin{figure*}[htb]
\centering
\includegraphics[width=0.99\textwidth]{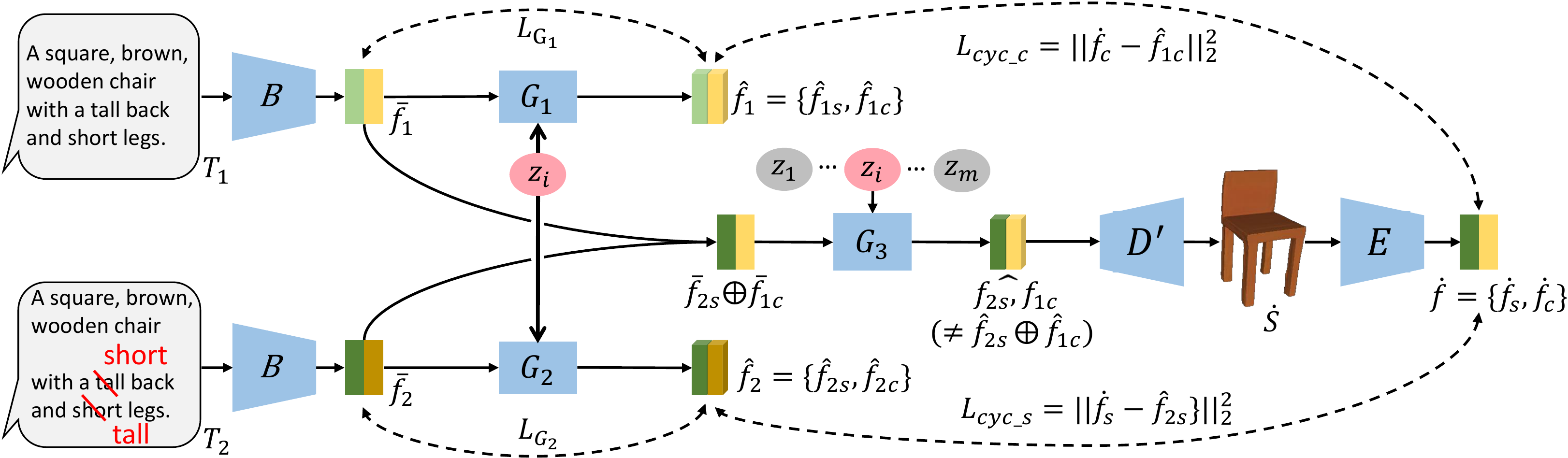}
\vspace*{-1mm}
\caption{Overview of our text-guided shape manipulation framework (with color unchanged). 
Given two pieces of text $\textbf{T}_1,\textbf{T}_2$, shape IMLE $G_1$ and $G_2$ use the same random noise $z_i$ to generate shapes. $G_3$ takes $\{\bar{f}_{2s},\bar{f}_{1c}\}$ and $z_i$ as input to generate shape $\dot{\textbf{S}}$ with feature $\{\dot{f}_{s},\dot{f}_{c}\}$ (encoded by $E$), such that $\dot{f}_{s}$ and $\dot{f}_{c}$ should be similar to $\hat{f}_{2s}$ and $\hat{f}_{1c}$, 
respectively.
Hence, we propose a two-way cyclic loss ($L_{cyc\_c}$ and $L_{cyc\_s}$) to encourage shape consistency between $\dot{\textbf{S}}$ and $\textbf{T}_2$, and color consistency between $\dot{\textbf{S}}$ and $\textbf{T}_1$. $G_1, G_2, G_3$ share the same weights.}
\label{fig:overview_mani_shape}
\end{figure*}

\begin{figure}[!t]
\centering
\includegraphics[width=0.99\columnwidth]{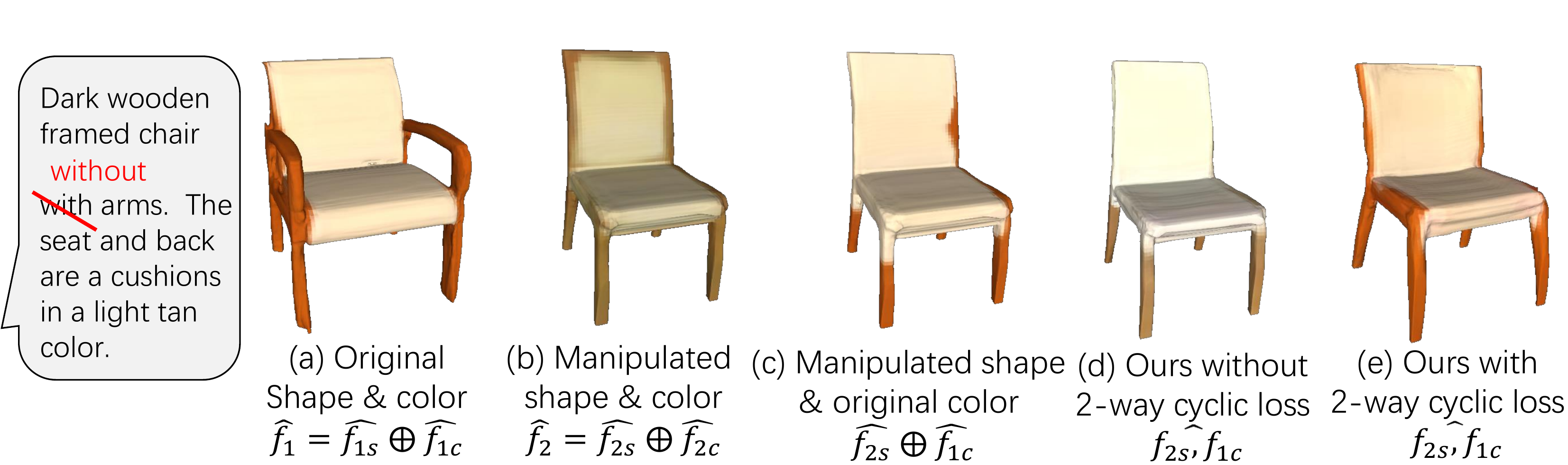}
\vspace*{-1mm}
\caption{(a) The original shape from the unedited text. (b) The shape from the edited text. It shows that even editing just a color-unrelated word may influence the generated color. (c) Replacing $\hat{f}_{2c}$ with the original color feature $\hat{f}_{1c}$ can cause misalignment between generated shape and color. (d) Our approach without the two-way cyclic loss, the unedited attributes may still change.
(e) Our full approach with the two-way cyclic loss produces an edited shape that better preserves the unedited attributes. 
}
\label{fig:2way-cyclic}
\end{figure}


\subsection{Text-Guided Shape Manipulation}

Next, we extend our framework for text-guided shape manipulation,~\ie, to generate shape $\dot{\mathbf{S}}$ that matches text $\mathbf{T_2}$ that is slightly modified from original text $\mathbf{T_1}$ by replacing/inserting/removing one or a few words, with other attributes unchanged for the same random noise $z$.

Taking shape manipulation (with color unchanged) as an example, we may directly feed feature $\hat{f}_{2} = \{\hat{f}_{2s},\hat{f}_{2c}\}$ from the edited text to $D^\prime$ to generate the new edited shape. Yet, it could cause drastic changes in the unedited region and colors (Figure~\ref{fig:2way-cyclic}(b)). Considering the decoupled shape and color features, we may
mix $\hat{f}_{2s}$ from edited text and $\hat{f}_{1c}$ from original text as input to $D^\prime$.
This simple approach ensures the consistency of the unedited attributes but shape and color may not well align with the edited shape (Figure~\ref{fig:2way-cyclic}(c)), since $\hat{f}_{2s}$ and $\hat{f}_{1c}$ actually come from different texts.
%

To encourage shape-color alignment, we propose to feed shape feature $\bar{f}_{2s}$ (extracted
from text $\mathbf{T}_2$) and color feature $\bar{f}_{1c}$ (extracted from text $\mathbf{T}_1$) to $G_3$ to predict the manipulated feature $\hat{{f}_{2s},{f}_{1c}}$.
%
Then, we can feed $\hat{{f}_{2s},{f}_{1c}}$ to $D^\prime$ to produce the edited shape $\dot{S}$.
Yet, this approach could still lead to certain changes in the unedited attributes (Figure~\ref{fig:2way-cyclic}(d)).
Figure~\ref{fig:overview_mani_shape} shows our full framework further with
%
the two-way cyclic loss,~\ie, $L_{cyc\_c}$ and $L_{cyc\_s}$.
Here, we use shape encoder $E$ to extract manipulated feature $\dot{f}=\{\dot{f_s},\dot{f_c}\}$ from $\dot{S}$ and formulate 
$L_{cyc\_s}$ for shape consistency ($\dot{f}_s$ and $\hat{f}_{2s}$) and 
$L_{cyc\_c}$ for color consistency ($\dot{f}_c$ and $\hat{f}_{1c}$).
Then, we can formulate the overall loss:
%
%
\begin{equation}
\begin{aligned}
L_{mani}&=(||\dot{f}_s-\hat{f}_{2s}||^2_2+||\dot{f}_c-\hat{f}_{1c}||^2_2)\mathbbm{1}(\textnormal{IoU}(I_1,I_2)>t)\\
&+L_{G_1} + L_{G_2},
\label{equ:2-way-cyclic}
\end{aligned}
\end{equation}
%
where the first term is the two-way cyclic consistency loss, which takes effect only when the Intersection over Union (IoU) between the associated ground-truth shapes $I_1$ and $I_2$ is larger than threshold $t$.
The last two terms fine-tune the shape IMLE for a diversified generation (see Eq.~\eqref{equ:imle}).

To train the framework, we initialize its weights from shape IMLE then finetune $G$ using $L_{mani}$ with all other modules $E,B,D^\prime$ frozen.
Also, we randomly sample two unpaired texts $\mathbf{T}_1$, $\mathbf{T}_2$ to simulate the original and edited texts.
With the two-way cyclic loss, the shape IMLE can learn to generate edited shapes with other attributes unchanged, while better aligning the shape and color.
Please see the supplementary material for the details on the color manipulation framework.
Besides Figures~\ref{fig:demo}(b,c) and~\ref{fig:2way-cyclic}(a,e), Figure~\ref{fig:manipulate_sec3} shows two more text-guided manipulation results.






\section{Experiments}

\subsection{Dataset and Implementation Details}
Our approach is evaluated on the largest text-shape dataset \textit{ShapeNet 3D models with natural language descriptions}~\cite{chen2018text2shape}.
The dataset contains $15,038$ shapes from the table and chair classes of ShapeNet~\cite{shapenet2015}; $75,344$ natural language descriptions, $16.3$ words per description on average, and $8,147$ unique words in the whole dataset~\cite{chen2018text2shape}.
%

\begin{figure}[!t]
\centering
\includegraphics[width=0.99\columnwidth]{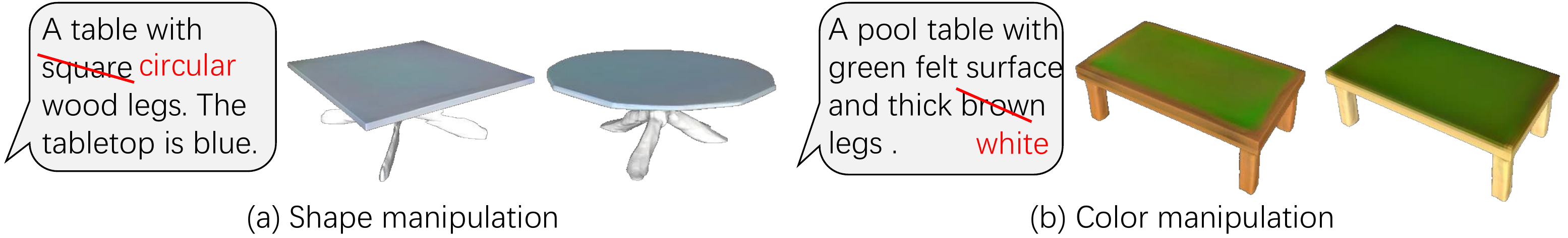}
\vspace*{-1mm}
\caption{Our text-guided shape and color manipulation results.}
\label{fig:manipulate_sec3}
\vspace*{-1mm}
\end{figure}


We implement our framework in PyTorch~\cite{paszke2019pytorch}.
To train the shape auto-encoder, we sample $4,096$ points with the strategy in~\cite{chen2019learning} and train the network for $500$ epochs in $16^3$ 
resolution, then continue the training for another $500$ epochs in $32^3$ 
resolution with learning rate $1e^{-4}$.
For text-guided shape generation, we train the network end-to-end for $200$ epochs, 
then fine-tune it end-to-end in $64^3$ 
resolution for another $200$ epochs.
For diversified shape generation, we train the shape IMLE for $100$ epochs with learning rate $1e^{-3}$ and the other network modules frozen.
Lastly, we fine-tune the shape IMLE for another $100$ epochs with the two-way cyclic consistency loss 
to enable manipulation.
We set hyper-parameters $d$, $d_l$, $\lambda_s$, $\lambda_c$, $\lambda_{reg}$, $\lambda_{cyc}$, and $t$ as $256$, $32$, $2$, $1$, $1$, $0.005$, and $0.01$, respectively, using a small validation set.

\subsection{Comparison with the Existing Works}

We compare our method with two existing works~\cite{chen2018text2shape,jahan2021semantics} (see also Section~\ref{sec:RW}) on text-guided shape generation.

\begin{figure}
\centering
\includegraphics[width=0.99\columnwidth]{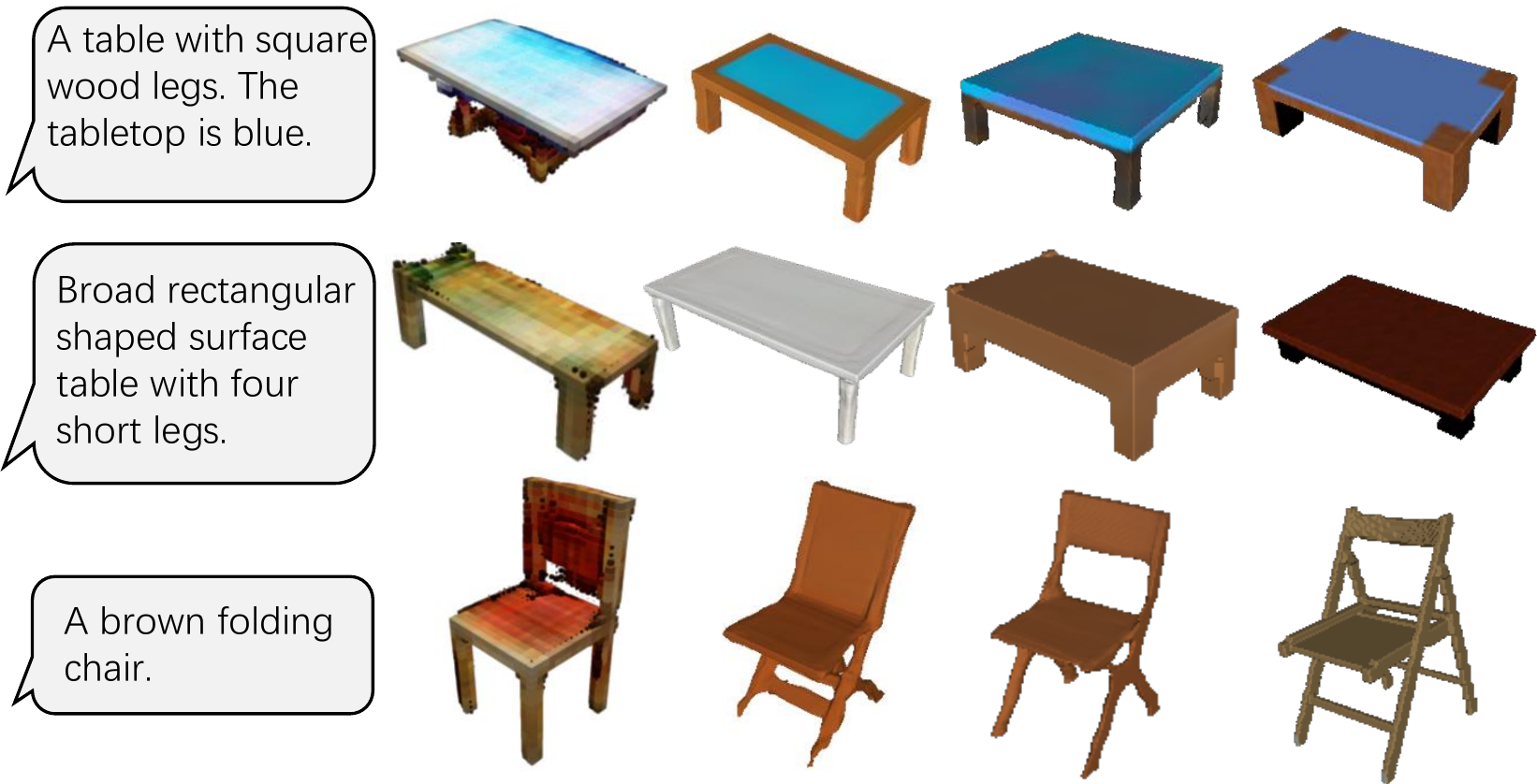}
\vspace*{-1mm}
\caption{Results by Text2shape~\cite{chen2018text2shape} (a) \vs ours (b,c) \vs GT (d). }
\label{fig:text2shape}
\vspace*{-1mm}
\end{figure}

For a fair comparison with~\cite{chen2018text2shape}, we transform our generated results into voxels in the same resolution as~\cite{chen2018text2shape},~\ie, $32^3$.
Also, we follow its train/val/test ($80\%/10\%/10\%$) split and its evaluation metrics,~\ie, IoU, EMD, IS, and Acc (Err=1-Acc), and directly compare our results with the numbers in~\cite{chen2018text2shape}.
Table~\ref{tab:existing} reports the results, showing that our method outperforms~\cite{chen2018text2shape} for all evaluation metrics, manifesting its effectiveness.
Note that ``IS'' ranges $[0,2]$, as it is built upon a two-category classification model, so both methods (1.96 {\vs} 1.97) already achieve satisfying performance in this respective.
The qualitative comparisons in Figure~\ref{fig:text2shape} also demonstrate the superiority of our approach, which is able to generate much better shapes and colors (see Figure~\ref{fig:text2shape} (b, c)) in comparison with~\cite{chen2018text2shape} (see Figure~\ref{fig:text2shape} (a)).

\begin{figure}
\centering
\includegraphics[width=0.95\columnwidth]{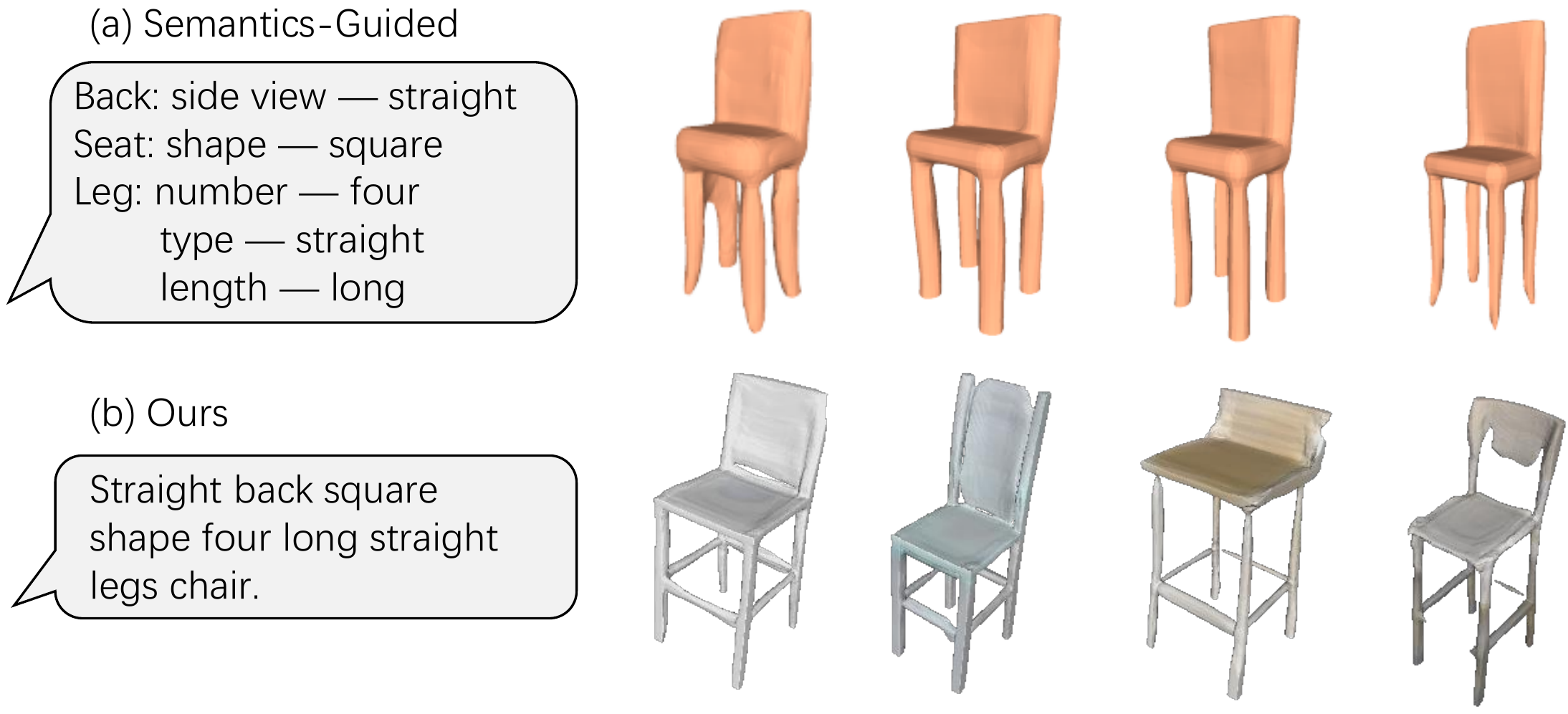}
\vspace*{-1mm}
\caption{Results generated by~\cite{jahan2021semantics} (a) \vs ours (b).}
\label{fig:semantic-guided}
\vspace*{-1mm}
\end{figure}

The other work~\cite{jahan2021semantics} focuses on generating shapes from phrase descriptions; see the left side of Figure~\ref{fig:semantic-guided} (a).
Since its setting is very different from ours, we only compare with it qualitatively.
To do so, we first prepare sentence descriptions that match the phrase descriptions in~\cite{jahan2021semantics} and then use our model to generate 3D results.
Comparing the results shown in Figures~\ref{fig:semantic-guided} (a) and (b), we can see that
our model is able to generate more diverse chairs that match the input description (``square shape, long straight leg''), while having varying colors and higher fidelity;
please see also the supplementary material for more comparison results.



\begin{table}
\centering
\caption{Quantitative comparisons with the existing work~\cite{chen2018text2shape}.
}
\vspace*{-1mm}
\scalebox{0.88}{
\centering
  \begin{tabular}{c|cccc}
    \toprule
    Method & IoU ($\uparrow$) & IS ($\uparrow$) & EMD ($\downarrow$) & Err ($\downarrow$) \\ 
    \midrule
    Text2Shape~\cite{chen2018text2shape} & 9.64 & 1.96 & 0.4443 & 2.63\\ 
    Ours & ~\textbf{12.21} & ~\textbf{1.97} & ~\textbf{0.2071} & ~\textbf{2.52} \\ 
    \bottomrule
  \end{tabular}
  }
\label{tab:existing}
\end{table}

\subsection{Ablation Studies}

We conduct extensive ablation studies to validate the effectiveness of the key components in ``text-guided shape generation'' and ``diversified generation.''
To measure the diversity and quality of the generated shapes, we formulate two new metrics, PS and FPD, based on Inception Score (IS)~\cite{salimans2016improved} and Fr$\acute{\text{e}}$chet Inception Distance (FID)~\cite{heusel2017gans}; please see the supplementary material for the details.
To evaluate the text-shape consistency, we adopt R-Precision~\cite{xu2018attngan}.
%
To reduce the training time, we train all models in $32^3$ resolution.

\vspace*{-3mm}
\paragraph{Text-guided shape generation.}
We evaluate the effectiveness of the following major components in this module (Figure~\ref{fig:overview} (a,b)): 
joint training with a pre-trained auto-encoder (AE),
decoupled shape-color decoder (DSCD),
WLST module (WLST), and
cyclic loss (CL).
Please refer to the supplementary material for the details of each setup.
%

\begin{figure}
\centering
\includegraphics[width=0.99\columnwidth]{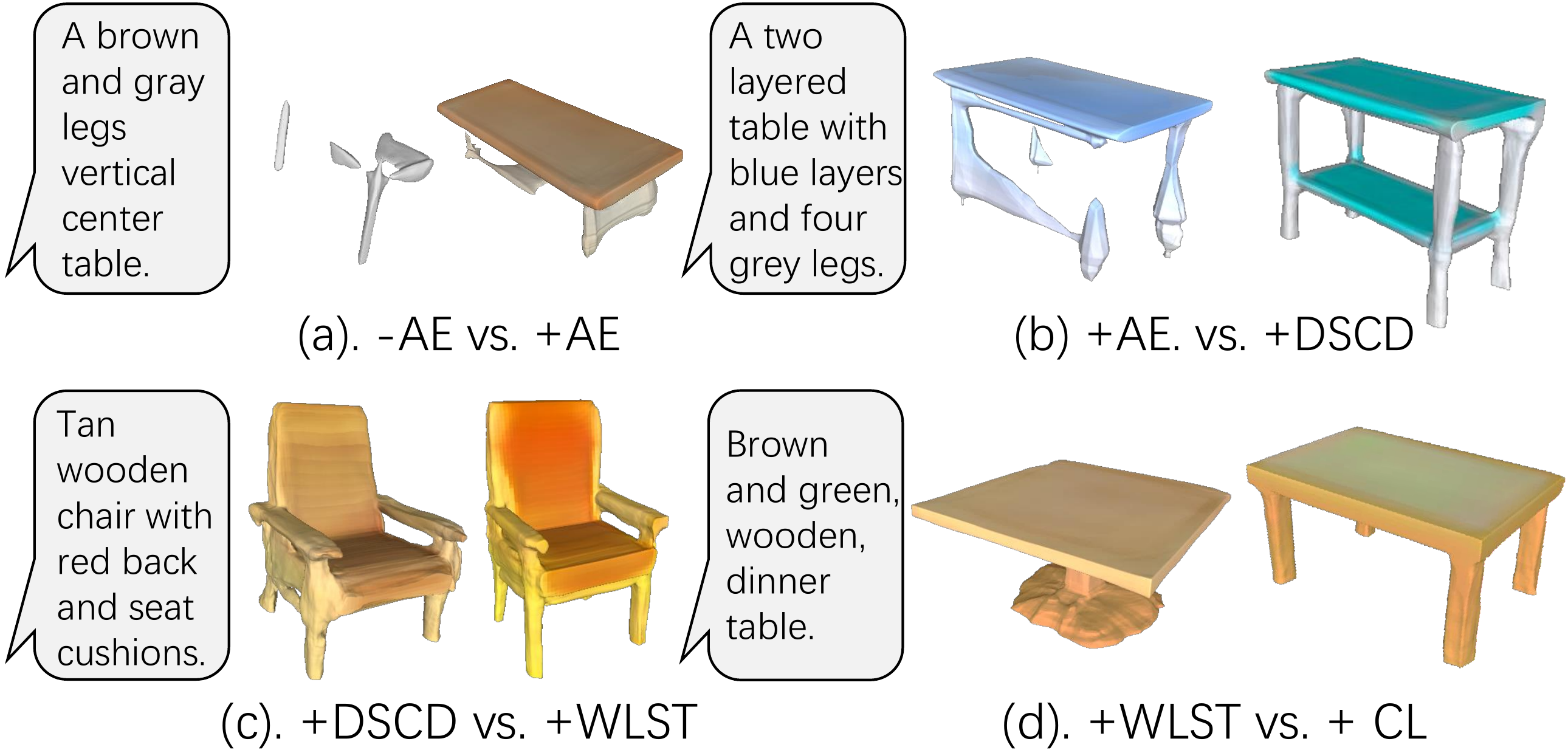}
\vspace*{-1mm}
\caption{Qualitative ablation studies on shape generation.}
\label{fig:ablation}
\end{figure}

Quantitative and qualitative results are shown in Table~\ref{tab:one-to-one} and Figure~\ref{fig:ablation}, respectively.
Note that all models in this setting achieve satisfying R-Precision ($>98\%$ except ``Without AE''), so we report R-Precision only in the next ``Diversified generation'' setting.
First, auto-encoder joint training (AE) is crucial for model convergence. Without AE, the baseline approach fails to converge, leading to unreasonable results (see Figure~\ref{fig:ablation} (a)) of very low quality. 
Second, decoupling shape and color in the decoder structure (DSCD) improves both PS and FPD by a large margin, manifesting its effectiveness in promoting high-fidelity and diversified synthesis.
This is also verified in the qualitative comparison shown in Figure~\ref{fig:ablation} (b).
Third, empowered by the word-level correlation, we can enrich the local details; see ``red back and seat cushion'' in Figure~\ref{fig:ablation} (c).
%
Lastly, cyclic loss (CL) improves the consistency between the generated shape and input text; see the visual comparison in Figure~\ref{fig:ablation} (d).
Note that both WLST and CL benefit the model more in the ``Diversified generation'' component to be detailed below.

\begin{table}
\centering
\caption{Ablation studies on text-guided shape generation.}
\vspace*{-1mm}
\label{tab:one-to-one}
\scalebox{0.9}{
  \begin{tabular}{c|ccccc}
    \toprule
    Method & IoU ($\uparrow$) & PS ($\uparrow$) & FPD ($\downarrow$)\\
    \midrule
    Without AE & 0.03 & 1.01$\pm$0.00 & 67.37\\
     \midrule
    +AE& 12.04 & 2.95$\pm$0.03 & 35.05\\
    further +DSCD& 12.00  & 3.16$\pm$0.04 &	31.09 \\
    further +WLST & 12.24  &3.21$\pm$0.05 & \textbf{30.34}\\
    further +CL (full) & \textbf{12.33}  &\textbf{3.26}$\pm$\textbf{0.06} & 30.80\\
    \bottomrule
  \end{tabular}
}
\end{table}


\begin{table}
\centering
\caption{Ablation studies on diversified shape generation.
%
}
\label{tab:multi-modal}
\vspace*{-1mm}
\scalebox{0.85}{
  \begin{tabular}{c|ccc}
    \toprule
    Method & PS ($\uparrow$) & FPD ($\downarrow$) & R-Precision ($\uparrow$)\\
    \midrule
    Latent GAN & 3.31 $\pm$ 0.02	&  30.70 &	21.20 $\pm$ 0.11  \\
    \midrule
    FC IMLE & 2.93 $\pm$ 0.02 &	29.53 & 25.97 $\pm$ 0.09\\
    Shape IMLE & 3.39 $\pm$ 0.02 &	29.65 & 27.60 $\pm$ 0.39 \\
    \midrule
    further +WLST & 3.39 $\pm$ 0.03 & 28.41 & 34.37 $\pm$ 0.09 \\
    +WLST+CL (full) & \textbf{3.45 $\pm$ 0.02} &	\textbf{27.26}	& \textbf{40.71} $\pm$ \textbf{0.10} \\
    \bottomrule
  \end{tabular}
  }
\end{table}

\begin{figure}
\centering
\includegraphics[width=0.99\columnwidth]{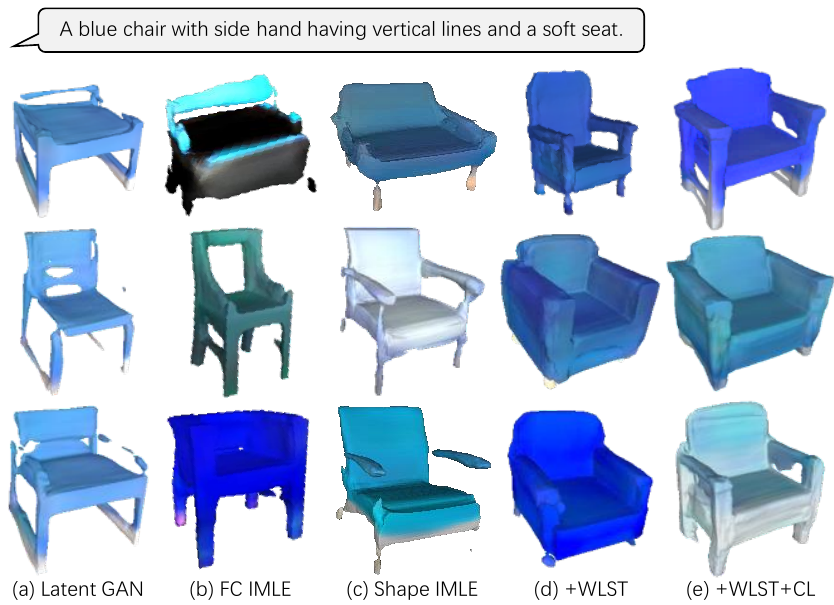}
\caption{Qualitative ablation studies on diversified generation.}
\label{fig:diverse}
\vspace*{-2mm}
\end{figure}

\vspace*{-3mm}
\paragraph{Diversified generation.} 
Next, we evaluate the major modules for diversified generation (Figure~\ref{fig:overview} (c)).
First, we replace the style-based shape IMLE with 
%
two different components for shape generation: a Latent GAN and a fully-connected IMLE (FC IMLE).
Besides, we explore models without the proposed WLST module and cyclic loss (CL), which benefit diversified shape generation.
Please refer to the supplementary material for the details of each setup.



Quantitative and qualitative results are shown in Table~\ref{tab:multi-modal} and Figure~\ref{fig:diverse}, respectively.
Note that this setting focuses on shape diversity and quality, so we do not adopt ``IoU,'' which measures the shape similarity to the ground truths.
In comparison with ``Latent GAN,'' the IMLE model can synthesize diversified colors and avoid generating collapsed invalid shapes (see Figure~\ref{fig:diverse} (a) {\vs} (b)), while attaining better quantitative results.
Further, the proposed style-based generator (``shape IMLE'') consistently improves on all metrics and yields higher quality shapes with better completeness 
in comparison with ``FC IMLE,'' as shown in Figure~\ref{fig:diverse} (c). 
Lastly, the WLST module and cyclic loss further help improves the generation fidelity and text-shape consistency by a large margin as shown in the last two rows of Table~\ref{tab:multi-modal}, manifesting their effectiveness (see Figure~\ref{fig:diverse} (d,e)).


\vspace{-1mm}
\subsection{Text-Guided Shape and Color Manipulation}

\begin{figure}
\centering
\includegraphics[width=0.999\columnwidth]{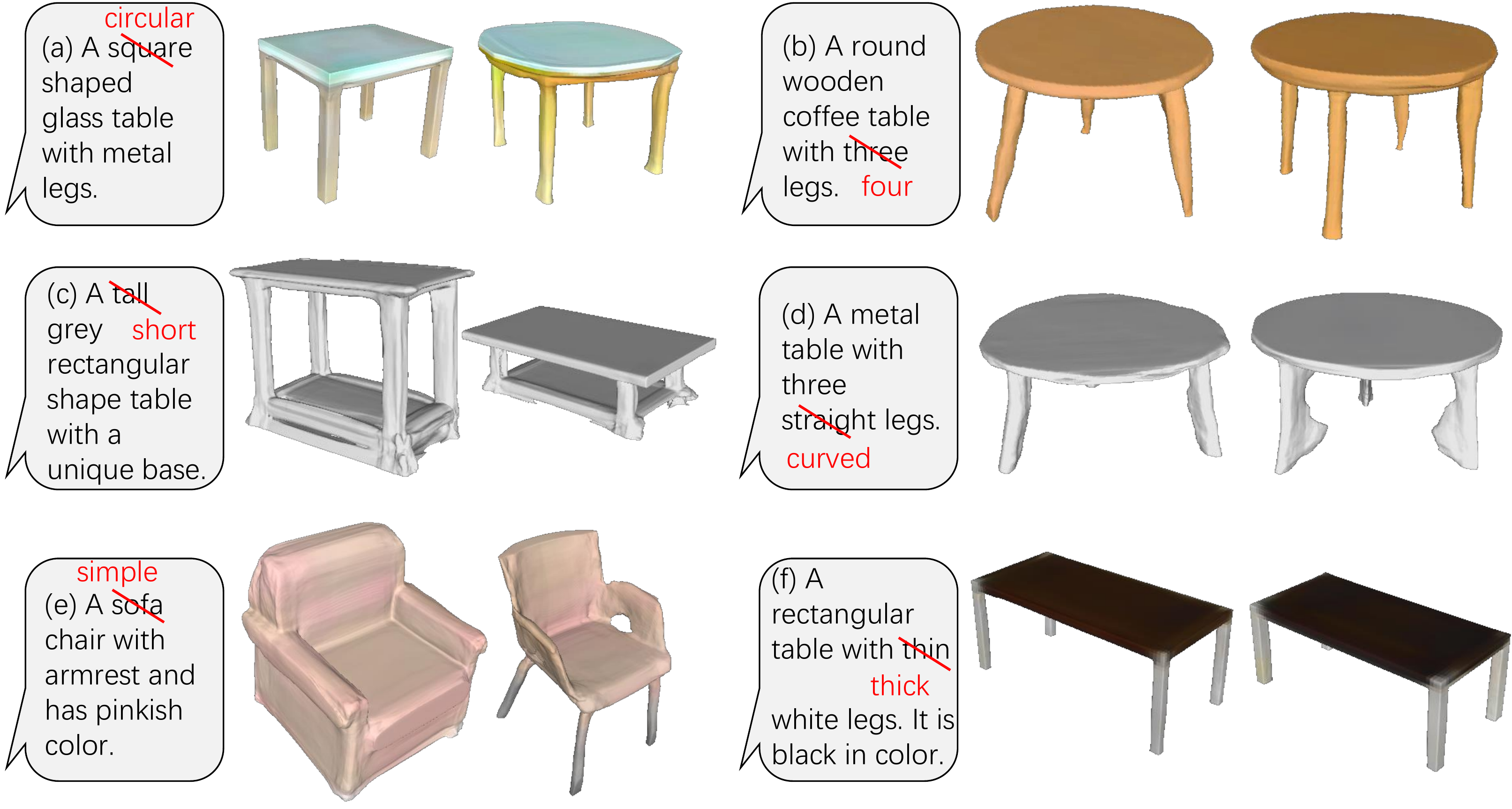}
\caption{Our text-guided shape manipulation results. We can manipulate (a) the shape of a table, (b) number of legs, (c) height, (d) shape of legs, (e) structure, (f) thickness of legs, and so on.
}
\label{fig:manipulate_shape}
\end{figure}
\begin{figure}
\centering
\includegraphics[width=0.999\columnwidth]{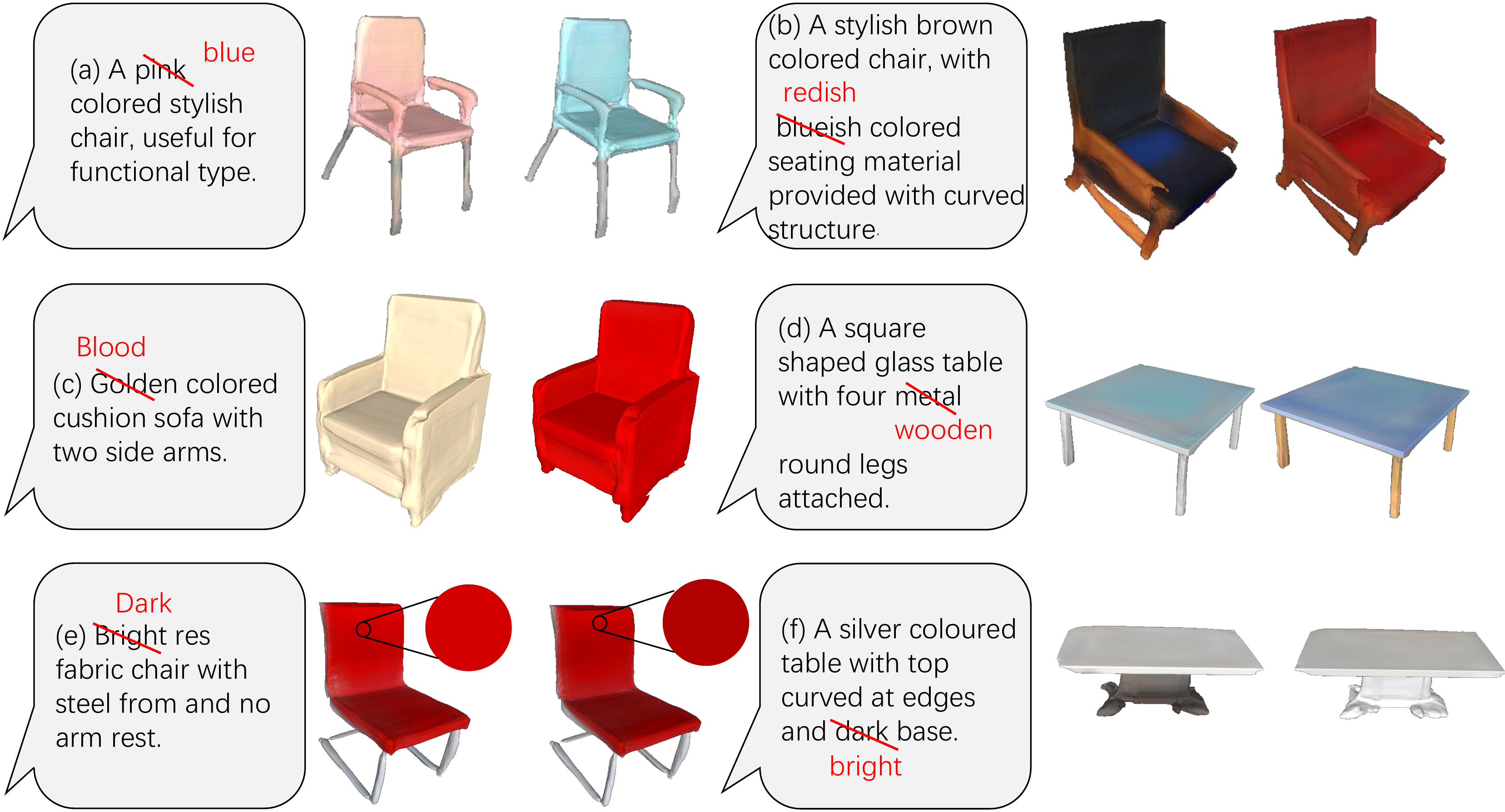}
\caption{Our text-guided color manipulation results. We can (c) manipulate the color indirectly using the related word (\eg, ``blood'' for red), (d) manipulate the material, (e,f) adjust the color brightness using words such as ``bright'' and ``dark'',~\etc}
\label{fig:manipulate_color}
\end{figure}

More text-guided manipulation results are shown in Figures~\ref{fig:manipulate_shape} and~\ref{fig:manipulate_color}, in addition to Figures~\ref{fig:demo}(b,c),~\ref{fig:2way-cyclic}(a,e), and~\ref{fig:manipulate_sec3}.
Thanks to our two-way cyclic loss, our model enables text-guided modification of colors and shapes in the generated results, while trying to keep the other attributes intact.
For instance, we are able to modify a ``square'' table to become ``circular,'' while keeping the other irrelevant regions unchanged,~\eg, the legs of the chair; see Figure~\ref{fig:manipulate_shape} (a). 
If we change the word ``pink'' to ``blue,'' 
only the associated parts in the shape are changed accordingly; see Figure~\ref{fig:manipulate_color} (a). 
More comparisons with the existing work~\cite{chen2018text2shape} and further ablation study on manipulation can be found in the supplementary material.

\vspace*{-3mm}
\section{Conclusion}
We have presented a novel framework capable of generating diversified 3D shapes with colors from text descriptions, while allowing flexible text-guided manipulations.
Besides the framework, we propose to decouple the shape and color predictions for learning both shape and color features from texts and design the word-level spatial transformer to explicitly correlate words with spatial locations to enhance the local details.
Also, we develop the cyclic consistency loss to enhance the text-shape consistency and introduce the style-based shape-IMLE generator for diversifying the generated shapes.
Further, we extend the framework for text-guided shape manipulation with the novel two-way cyclic loss.
Extensive experimental studies manifest the effectiveness of our framework.
Limitation analysis and future works are elaborated in the supplementary material.
\vspace*{-5mm}
\paragraph{Acknowledgement.}
This work is supported by the Research Grands Council of the Hong Kong Special Administrative Region (Project No. 14201921 and 27209621).

\clearpage
\appendix
\centerline{\Large{\textbf{Supplementary Material}}}

\section{Evaluation Metrics}

This section introduces the evaluation metrics employed in the experiments.
Below, we first introduce the metrics we formulated/extended from the existing ones for the evaluations, and then introduce other metrics from Text2Shape~\cite{chen2018text2shape}.

\begin{itemize}

\item{\textbf{PS} and \textbf{FPD}}

Point Score (PS) and Fréchet Point Distance (FPD) measure shape diversity and quality.

Existing works~\cite{shu20193d,li2021sp} often utilize the Fréchet Point Distance (FPD)
to evaluate the quality of the generated 3D shapes.  However, such metric cannot account for color, which is one of the important characteristics in our results that is not present in the previous works.  To jointly evaluate the shape and color in the generated results, we formulate PS and extend FPD for shape and color evaluations based on
Inception Score (IS)~\cite{szegedy2016rethinking} and Fréchet Inception Distance (FID)~\cite{heusel2017gans}. 


PS measures the KL-divergence between the conditional probabilities of the generated shapes and their marginal probabilities. 
On the other hand, FPD measures the Wasserstein distance between the distribution of the generated shapes and that of the real samples. 
The mentioned probabilities are inferred from a pre-trained classification network (\eg, Inception v3~\cite{szegedy2016rethinking} on ImageNet for image generation). In our case, PS and FPD are built upon a newly-trained PointNet~\cite{qi2017pointnet}, since no existing 3D classification network can simultaneously consider both shape and color as far as we know. Specifically, we train a classification-based PointNet on ScanObjectNN~\cite{uy2019revisiting} for 200 epochs, with a validation classification accuracy of 84.85\%.

\vspace*{-1mm}
\item{\textbf{IoU}}

Intersection over Union (IoU) measures the similarity to the ground truth.
We evaluate the IoU between the generated shape and ground truth, by measuring the similarity of the occupancy between them.

\item{\textbf{R-precision}}\\
We adopt \textbf{R-precision}~\cite{xu2018attngan} to measure the consistency between the generated shape \textbf{S} and input text \textbf{T}. Specifically, we extract shape and text features using $E$ and $B$, respectively, then evaluate R-Precision three times with different random seeds to reduce the randomness.


\item{Metrics in Text2Shape~\cite{chen2018text2shape}}

Chen~\etal~\cite{chen2018text2shape} adopt four metrics, including \textbf{IoU}, \textbf{EMD}, \textbf{IS}, and \textbf{Acc} (\textbf{Err}=1-Acc), for evaluating their results.
IoU and EMD measure the shape and color similarity between the generated shape and ground truth, respectively. IS measures the diversity and quality of the generated shapes, and Err (Acc) measures the quality. Please refer to~\cite{chen2018text2shape} for the details.
For a pair comparison, we train a classification model using the official code released by the author of~\cite{chen2018text2shape} to evaluate our IS and Acc. 

\end{itemize}

\section{Details of the Baselines}
\subsection{Text-Guided Shape Generation}~\label{sec:baseline1}

We create the following baselines to evaluate the key modules of our text-guided shape generation framework, including the auto-encoder (AE), decoupled shape-color decoder (DSCD), WLST module (WLST), and cyclic loss (CL). 
\begin{itemize}
\vspace*{-1mm}
\item[(i)] ``Without AE.''
In this setting, the network is composed of text encoder $B$ and decoder $D$ but without shape encoder $E$. It is optimized to regress the the occupancy and color values of the target shape $I$ with an $L_2$ loss. 
It serves as our preliminary baseline for text-guided shape generation, where $B$ maps the text \textbf{T} to a latent space and $D$ reconstructs the shape and color.
\item[(ii)] ``+AE.''
The network is composed of the auto-encoder $E$, $D_{share}$, and text encoder $B$. $E$ encodes the input shape $I$ as a joint shape-color feature $f_{share}$ and $D$ reconstructs the shape and color of $I$. $B$ extracts the text feature $\bar{f}_{share}$ and minimizes the mean squared difference between $\bar{f}_{share}$ and $f_{share}$. It serves as a baseline to directly adopt the auto-encoder-based approach~\cite{chen2019learning} to our task, as introduced in Section 1 of the main paper, so this baseline demonstrates the necessity of the auto-encoder in our approach.  
\item[(iii)] ``Further +DSCD.''
In this setting, we replace the shared decoder $D_{share}$ with a pair of decoupled shape-color decoders $D=\{D_s,D_c\}$ and replace shared features $f_{share}$ and $\hat{f}_{share}$ in (ii) with a pair of decoupled features $f=\{f_s,f_c\}$ and $\hat{f}=\{\hat{f}_s,\hat{f}_c\}$, respectively. This baseline manifests the effectiveness of the decoupled shape-color decoder (DSCD) in our framework.
\item[(iv)] ``Further +WLST.''
Based on (iii), the spatial-aware decoder $D^\prime$ with the WLST is adopted for text-guided shape generation in place of $D$. This baseline manifests the effectiveness of WLST. 
\item[(v)] ``Further +CL'' (our full model). Model (iv) is trained with an additional cyclic loss (which is Eq.(5) in the main paper), whereas Model (v) is our full model.
Comparing between model (iv) and model (v) verifies the applicability of the cyclic loss. 
\end{itemize}

\subsection{Diversified Generation}~\label{sec:baseline2}

To evaluate the core modules for diversified shape generation, We compare our Shape IMLE with two other approaches: (i) Latent GAN and (ii) fully-connected IMLE (FC IMLE). Besides, we evaluate the performance gain of our proposed WLST and cyclic loss (CL) for diversified shape generation.  For each baseline, we generate three different samples for each text with random noises $z_1$ to $z_3$.

\begin{figure}
\centering
\includegraphics[width=0.99\columnwidth]{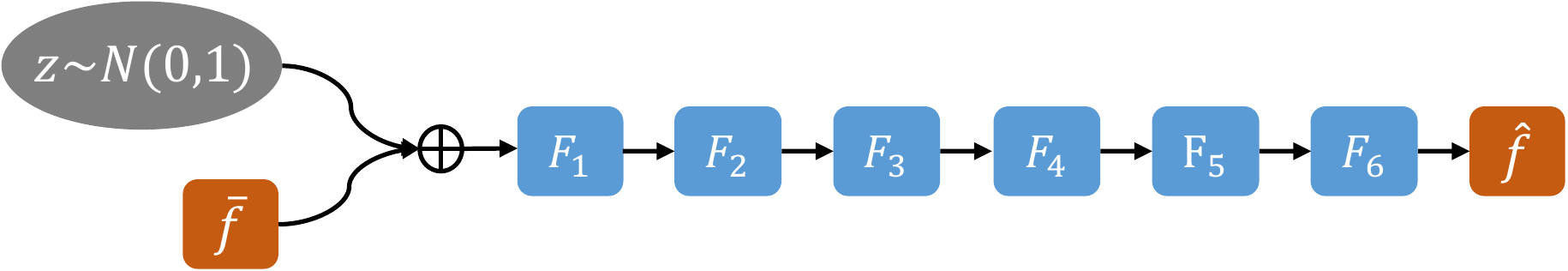}
\caption{The FC generator architecture in FC IMLE.}
\label{fig:simple}
\vspace*{-2mm}
\end{figure}

\begin{itemize}
\vspace*{-1mm}
\item[(i)] ``Latent-GAN.''
We adopt Latent-GAN~\cite{achlioptas2018learning2,arjovsky2017towards} conditioned on the input text to generate diversified results. We adopt our style-based latent shape-IMLE generator $G$ (Figure 5 in the main paper) as generator and a small network with three fully-connected layers as discriminator $D_\text{latent}$. We train the generator and discriminator iteratively using adversarial training with all the other modules frozen. 
\item[(ii)] ``FC IMLE.''
In this setting, we introduce the IMLE framework for diversified generation. As shown in Figure~\ref{fig:simple}, the IMLE generator $G_{simple}$ is composed of six fully-connected layers that take $\bar{f} \oplus z$ as input. This baseline aims to show the superiority of IMLE. 
\item[(iii)] ``Shape IMLE.''
In place of FC-IMLE in (ii), we adopt the style-based shape-IMLE generator $G$ shown in Figure 5 in the main paper. This baseline manifests the effectiveness of our shape-IMLE generator $G$. 
\item[(iv)] ``+WLST'' and ``+CL.''
Similar to ``+WLST'' and ``+CL'' in Section~\ref{sec:baseline1}, we again evaluate their capability on improving the diversified generation. 
\end{itemize}


\begin{figure}[!t]
\centering
\includegraphics[width=0.99\columnwidth]{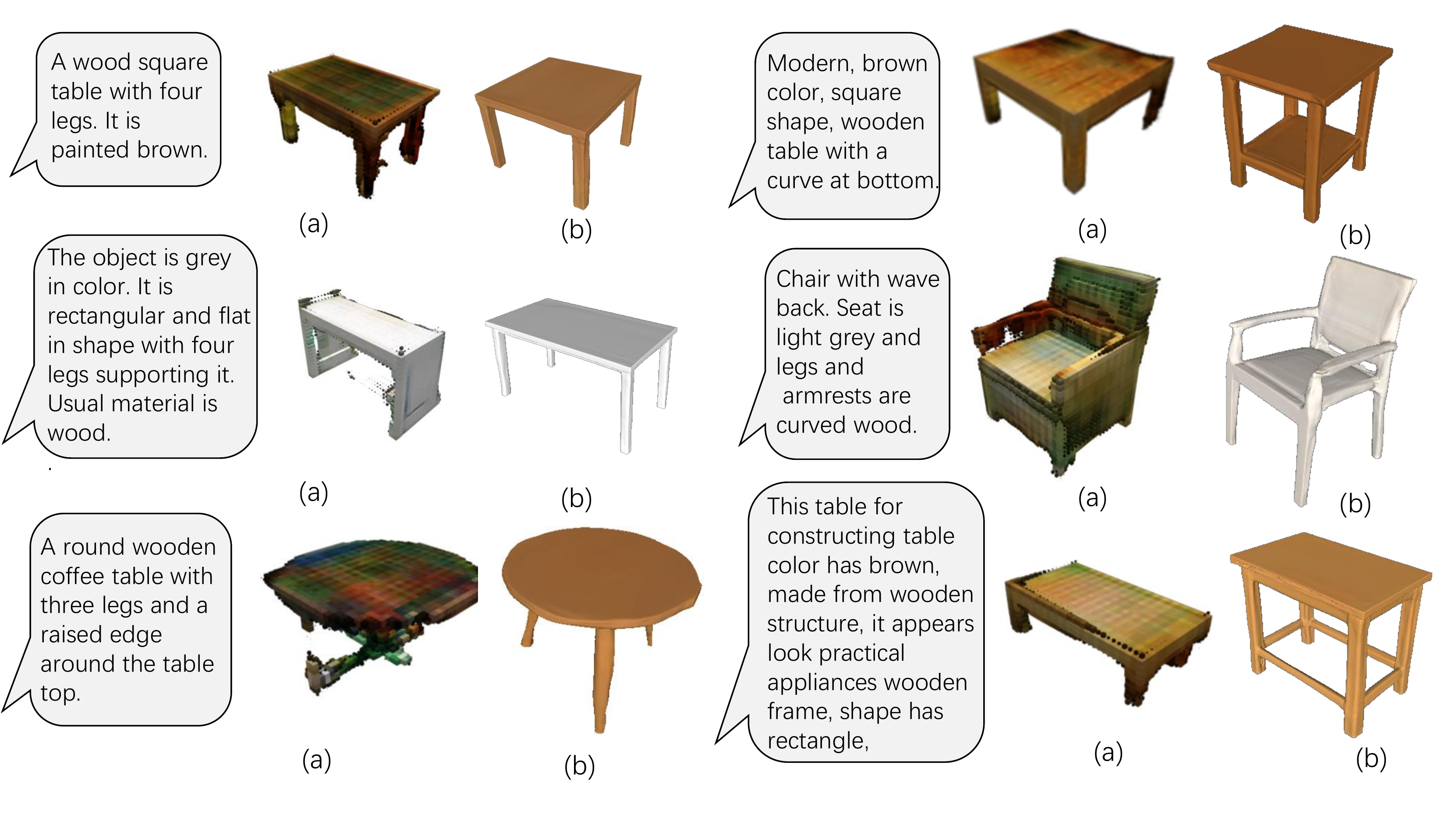}
\caption{Additional text-guided generation results compared with Text2Shape~\cite{chen2018text2shape}. }
\label{fig:text2shape2}
\vspace*{-2mm}
\end{figure}

\begin{figure}[!t]
\centering
\includegraphics[width=0.99\columnwidth]{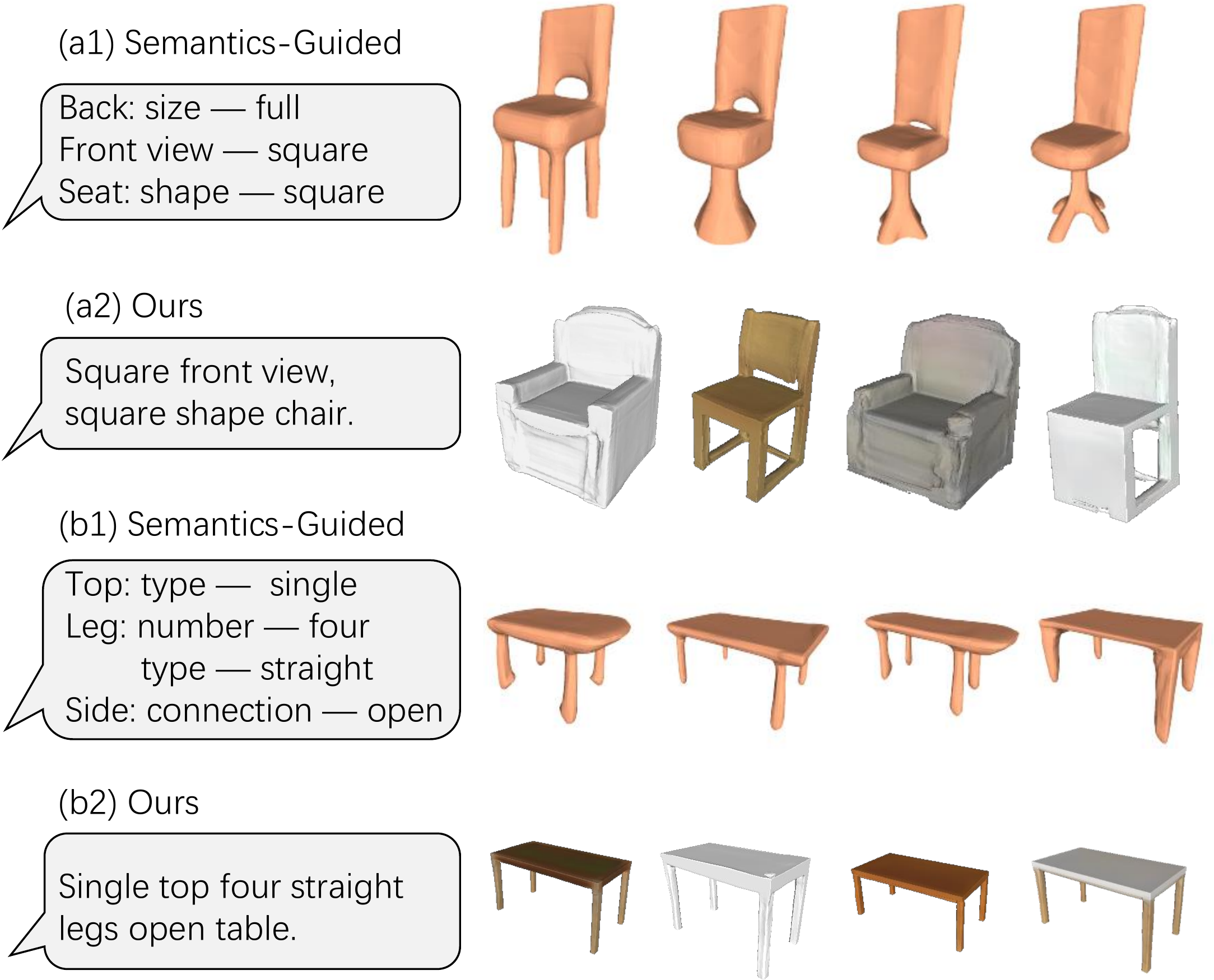}
\caption{Additional text-guided generation results compared with~\cite{jahan2021semantics}.}
\label{fig:semantics-guided2}
\vspace*{-2mm}
\end{figure}

\section{Results of Text-Guided Shape Generation}
\subsection{Comparison with Existing Works}

In this section, we show more results on comparing our method with~\cite{chen2018text2shape} and~\cite{jahan2021semantics}. As shown in Figure~\ref{fig:text2shape2}, our approach is able to generate shapes with higher fidelity compared with ~\cite{chen2018text2shape}. Also, our results are more consistent to the input texts. As shown in Figure~\ref{fig:text2shape2} (e) on the bottom left of the figure, our approach is able to create a folding chair following the text description, where~\cite{chen2018text2shape} can only output a regular chair. 

The most recent work~\cite{jahan2021semantics} takes only pre-defined semantic labels as inputs, unlike our approach, which can take natural language as inputs.
As shown in Figure~\ref{fig:semantics-guided2}, our approach can generate more diversified shapes that better match the input text description (“square shape, square view” in Figure~\ref{fig:semantics-guided2} (a1, a2)), compared with~\cite{jahan2021semantics}.

\subsection{Additional Generation Results}

\begin{figure}
\centering
\includegraphics[width=0.99\columnwidth]{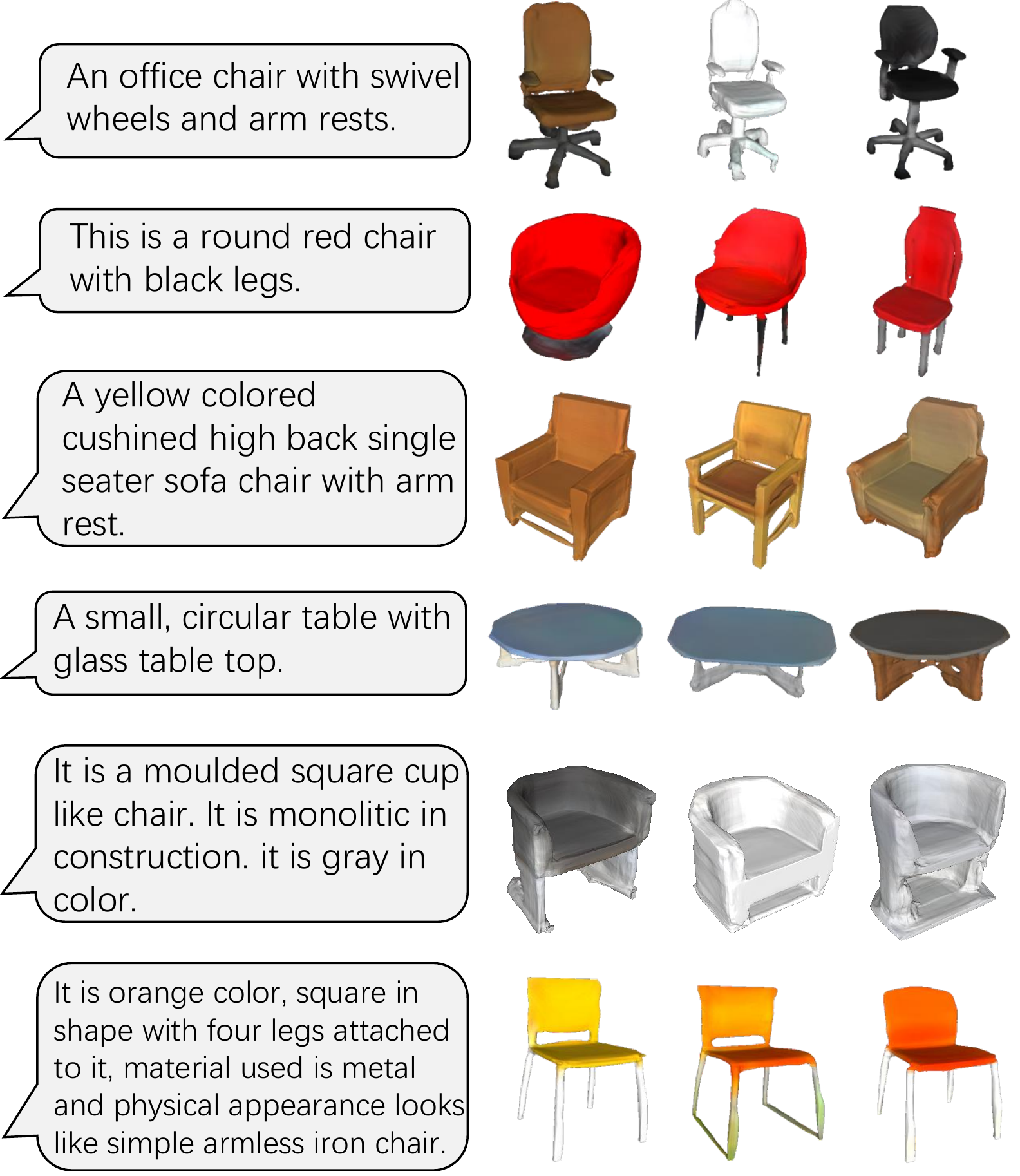}
\caption{Our text-guided shape generation results.}
\label{fig:more_div}
\vspace*{-2mm}
\end{figure}

Further, we show more text-guided shape generation results in Figures~\ref{fig:more_div}. These results again manifest the superiority of our approach on diversity, fidelity, and text-shape consistency, demonstrating the capability of our method over the previous ones.

\section{Text-Guided Shape Manipulation}

\subsection{Color Manipulation Framework}
\begin{figure*}[htb]
\centering
\includegraphics[width=0.99\textwidth]{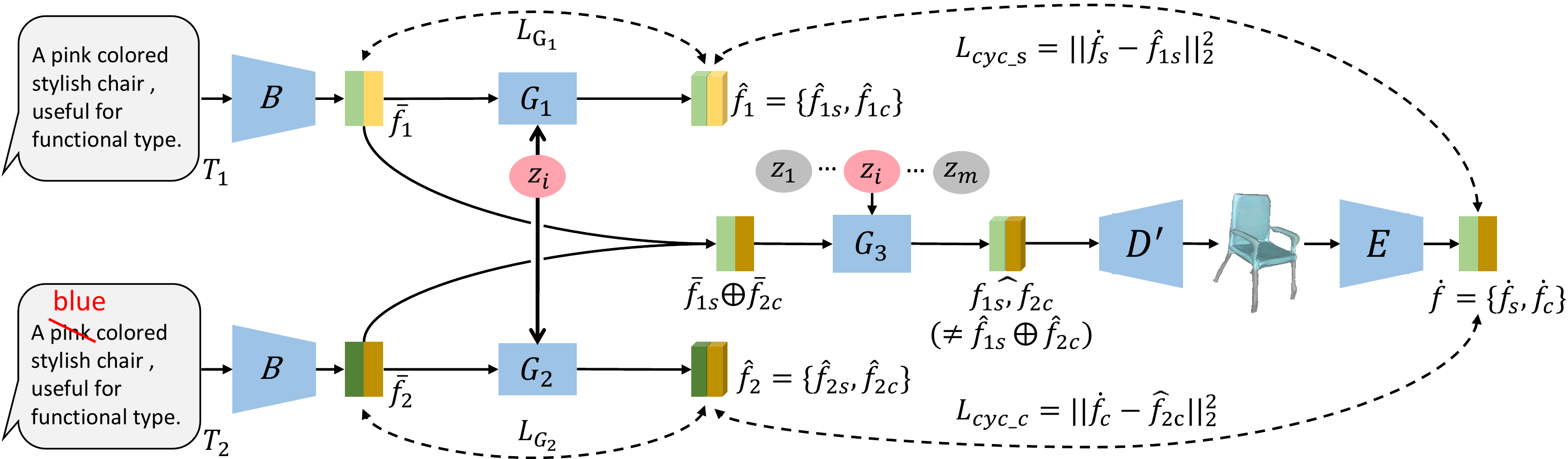}
\caption{Overview of our text-guided color manipulation framework (with shape unchanged). 
Given two pieces of text $\textbf{T}_1,\textbf{T}_2$, shape IMLE $G_1$ and $G_2$ use the same random noise $z_i$ for shape generation. $G_3$ takes $\{\bar{f}_{1s},\bar{f}_{2c}\}$ and $z_i$ as input to generate shape $\dot{\textbf{S}}$ with feature $\{\dot{f}_{s},\dot{f}_{c}\}$ (encoded by $E$), such that $\dot{f}_{s}$ and $\dot{f}_{c}$ should be similar to $\hat{f}_{1s}$ and $\hat{f}_{2c}$, respectively.
To this end, we propose a two-way cyclic loss to encourage the shape consistency between $\dot{\textbf{S}}$ and $\textbf{T}_1$, and the color consistency between $\dot{\textbf{S}}$ and $\textbf{T}_2$. $G_1, G_2, G_3$ share the same weights.
}
\label{fig:overview_mani_color}
\vspace*{-2mm}
\end{figure*}

In this section, we introduce our framework for text-guided color manipulation with shape unchanged. As shown in Figure~\ref{fig:overview_mani_color}, we feed shape feature $\bar{f}_{1s}$ (extracted from $\mathbf{T}_1$) and color feature $\bar{f}_{2c}$ (extracted from $\mathbf{T}_2$) to $G_3$ to predict the manipulated feature $\hat{{f}_{1s},{f}_{2c}}$,
%
then feed it to $D^\prime$ to produce the edited shape $\dot{S}$.
We then extract manipulated feature $\dot{f}=\{\dot{f_s},\dot{f_c}\}$ from $\dot{S}$ using $E$ and use the two-way cyclic loss,~\ie, $L_{cyc\_s}$ to encourage shape consistency ($\dot{f}_s$ and $\hat{f}_{1s}$) and $L_{cyc\_c}$ to encourage color consistency ($\dot{f}_c$ and $\hat{f}_{2c}$).

Similar to the shape manipulation framework shown in the main paper, we train our color manipulation framework using the following loss:
\begin{equation}
\begin{aligned}
L_{mani}^c&=(||\dot{f}_s-\hat{f}_{1s}||^2_2+||\dot{f}_c-\hat{f}_{2c}||^2_2)\mathbbm{1}(\textnormal{IoU}(I_1,I_2)>t)\\
&+L_{G_1} + L_{G_2},
\label{equ:2-way-cyclic-color}
\end{aligned}
\end{equation}
%
where the terms have the same definition as Eq.(8) in the main paper. 

\subsection{Comparison with the Existing Work}

\begin{figure}
\centering
\includegraphics[width=0.99\columnwidth]{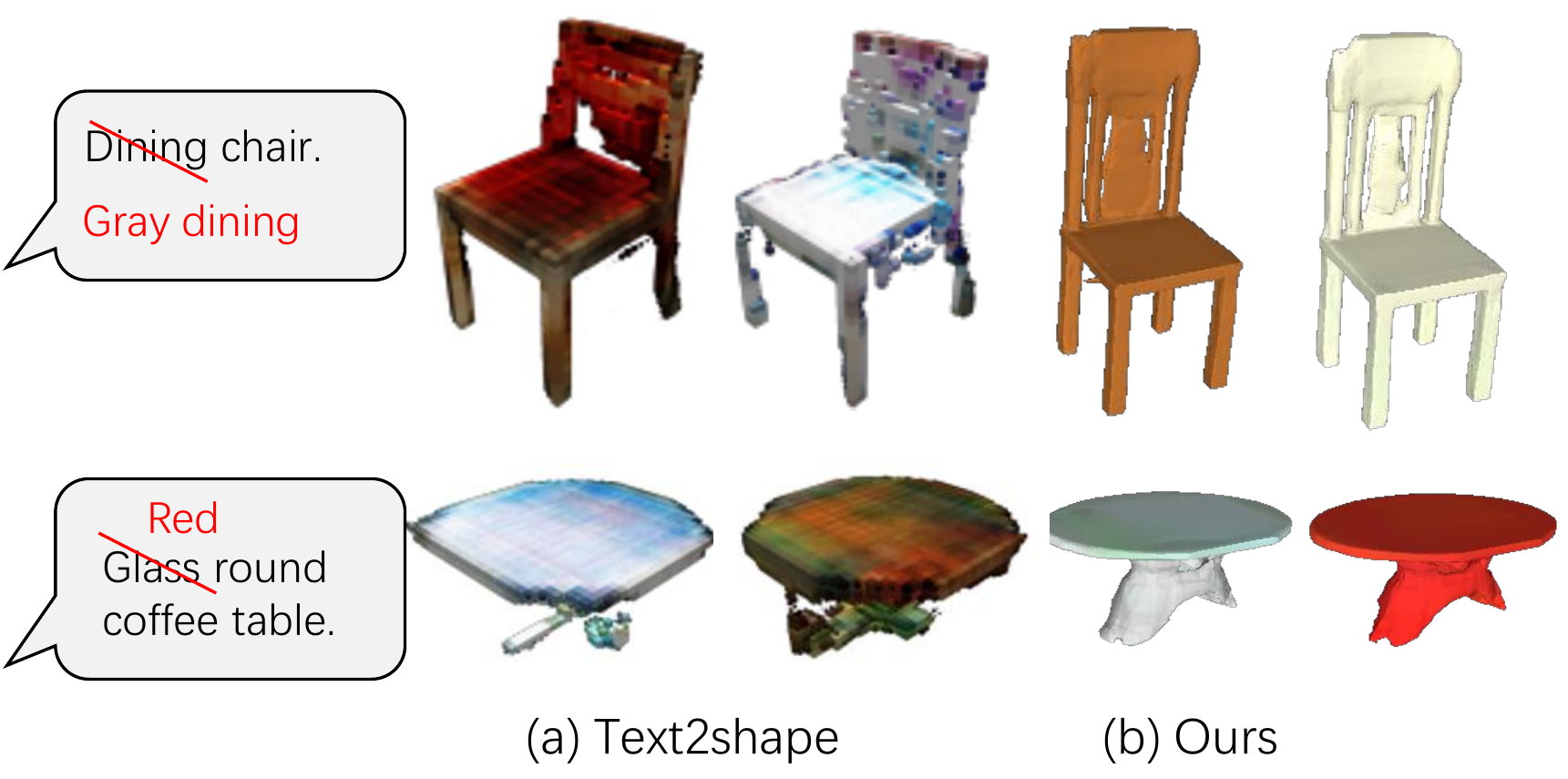}
\caption{Text-guided manipulation results on comparing our method with Text2Shape~\cite{chen2018text2shape}.}
\label{fig:manipulate_text2shape}
\vspace*{-2mm}
\end{figure}

In this section, we compare our method with~\cite{chen2018text2shape} on shape manipulation capability. As shown in Figure~\ref{fig:manipulate_text2shape}, inserting or editing words related to the color attribute leads to undesirable changes in the other attributes, as shown in the results produced by~\cite{chen2018text2shape},~\eg, the shape of the chair back and the table leg, whereas our approach is able to better preserve the shapes (geometries and structures).

\subsection{Ablation Studies}

In this section, we evaluate the different strategies for text-guided manipulation quantitatively. To measure the quality of the manipulated shapes, We adopt PS and FPD as the evaluate metrics. To further evaluate the consistency before and after the manipulation, we calculate R-Precision$_1$ based on $\dot{f}$ (feature from the manipulated shape) and $\hat{f}_1$ (feature from the original text), and assess R-Precision$_2$ based on $\dot{f}$ (feature from the manipulated shape) and $E(D^\prime(\hat{f}_1))$ (feature from the generated shape by the original text). We build a small dataset containing $50$ pairs of original and manipulated texts for the evaluation. 

\begin{itemize}
\vspace*{-1mm}
\item[(i)] Baseline 1. As shown in Figure 7(b) of the main paper, we directly feed the feature from the edited text $\hat{f}_2=\{\hat{f}_{2s},\hat{f}_{2c}\}$ to our generation framework. It is the primitive baseline for manipulation because it adopts no mechanism for consistency preserving, but it serves as the upper bound of the shape manipulation quality, since it directly adopts our generation framework (Figure 2 in the main paper) to produce the result without any constraints on the manipulation consistency.
\item[(ii)] Baseline 2. As shown in Figure 7(c) of the main paper, the shape is generated by a mixture of $\hat{f}_1$ and $\hat{f}_2$. Specifically, for the shape manipulation, we feed $\hat{f}_{2s} \oplus \hat{f}_{1c}$ to $D^\prime$; and for the color manipulation, we use $\hat{f}_{1s} \oplus \hat{f}_{2c}$. 
\item[(iii)] Baseline 3. As shown in Figure 7(d) of the main paper, we feed a mixture of $\bar{f}_1$ and $\bar{f}_2$ to $G$ to boost the shape-color alignment. For the shape manipulation, we feed $\bar{f}_{2s} \oplus \bar{f}_{1c}$ to $G$ to derive $\hat{f_{2s},f_{1c}}$; and for the color manipulation, we predict $\hat{f_{1s},f_{2c}}$ from $\hat{f}_{1s} \oplus \hat{f}_{2c}$ with $G$. 
\item[(iv)] Our full model (Ours). Built upon (iii), we further incorporate the two-way cyclic loss shown in Eq.(8) of the main paper for shape manipulation, and Eq.~\eqref{equ:2-way-cyclic-color} in this supplementary document for color manipulation. 
\end{itemize}

``Baseline 1'' generates a new shape using the edited text without considering what the original shape is. As shown in Table~\ref{tab:ablation_mani}, despite of the best diversity and quality it achieves, the lowest R-Precisions indicate the unsatisfying consistency before and after the manipulation (see Figure 7 (b) in the main paper). 

On the other hand, ``Baseline 2'' and ``Baseline 3'' attain better consistency at the expense of the generation quality, and our manipulation framework with the two-way cyclic loss is able to achieve the best consistency before and after the manipulation, while having better generation quality compared with both ``Baseline 2'' and ``Baseline 3,'' even being close to ``Baseline 1.''

\begin{table}
\centering
\caption{Ablation studies on text-guided manipulation.
}
\label{tab:ablation_mani}
\vspace*{-1mm}
\scalebox{0.75}{
  \begin{tabular}{c|cccc}
    \toprule
    Method & PS ($\uparrow$) & FPD ($\downarrow$) & R-Precision$_1$ ($\uparrow$) & R-Precision$_2$ ($\uparrow$)\\
    \midrule
    Baseline 1 & \textbf{2.80} $\pm$ \textbf{3.03} & \textbf{31.70} & 36.00 $\pm$ 2.83 & 48.67 $\pm$ 0.94 \\
    Baseline 2 & 2.73 $\pm$ 0.39 & 33.77 & 43.33 $\pm$ 0.94 & 52.00 $\pm$ 3.26\\
    Baseline 3 & 2.75 $\pm$ 0.48 & 35.74 & 42.66 $\pm$ 3.77 & 56.67 $\pm$ 2.49\\
    Ours & 2.76 $\pm$ 0.53 & 32.03 & \textbf{58.00} $\pm$ \textbf{2.82} & \textbf{67.33} $\pm$ \textbf{1.89}\\
    \bottomrule
  \end{tabular}
  }
\end{table}

\section{Alternative Training Strategy}
\begin{figure}
\centering
\includegraphics[width=0.99\columnwidth]{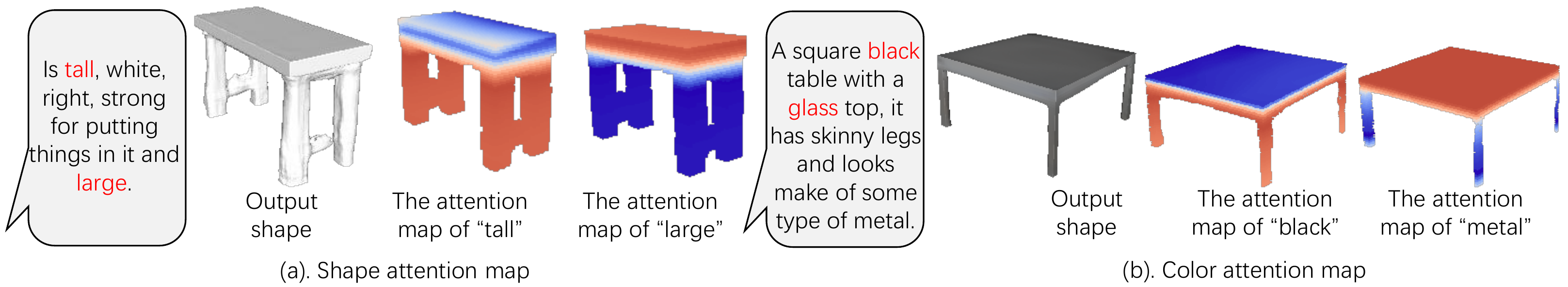}
\caption{Attention map of the end-to-end training strategy.}
\label{fig:attention2}
\vspace*{-2mm}
\end{figure}

In this section, we discuss an optional training strategy. Specifically, we jointly train the shape auto-encoder and text encoder $E,D^\prime$ and $B$ end-to-end, instead of following the training strategy presented in the main paper that first trains $E,D$, and then jointly trains $E,D^\prime,B$. This strategy includes fewer training steps, but needs much more training time because $B$ continuously optimized in the whole training process. This training strategy achieves comparable performance as presented in the main paper, and can generate attention maps that are more consistent with the semantic meaningful shape parts as shown in Figure~\ref{fig:attention2}. 

\section{Limitations and Future Work}
\begin{figure}
\centering
\includegraphics[width=0.99\columnwidth]{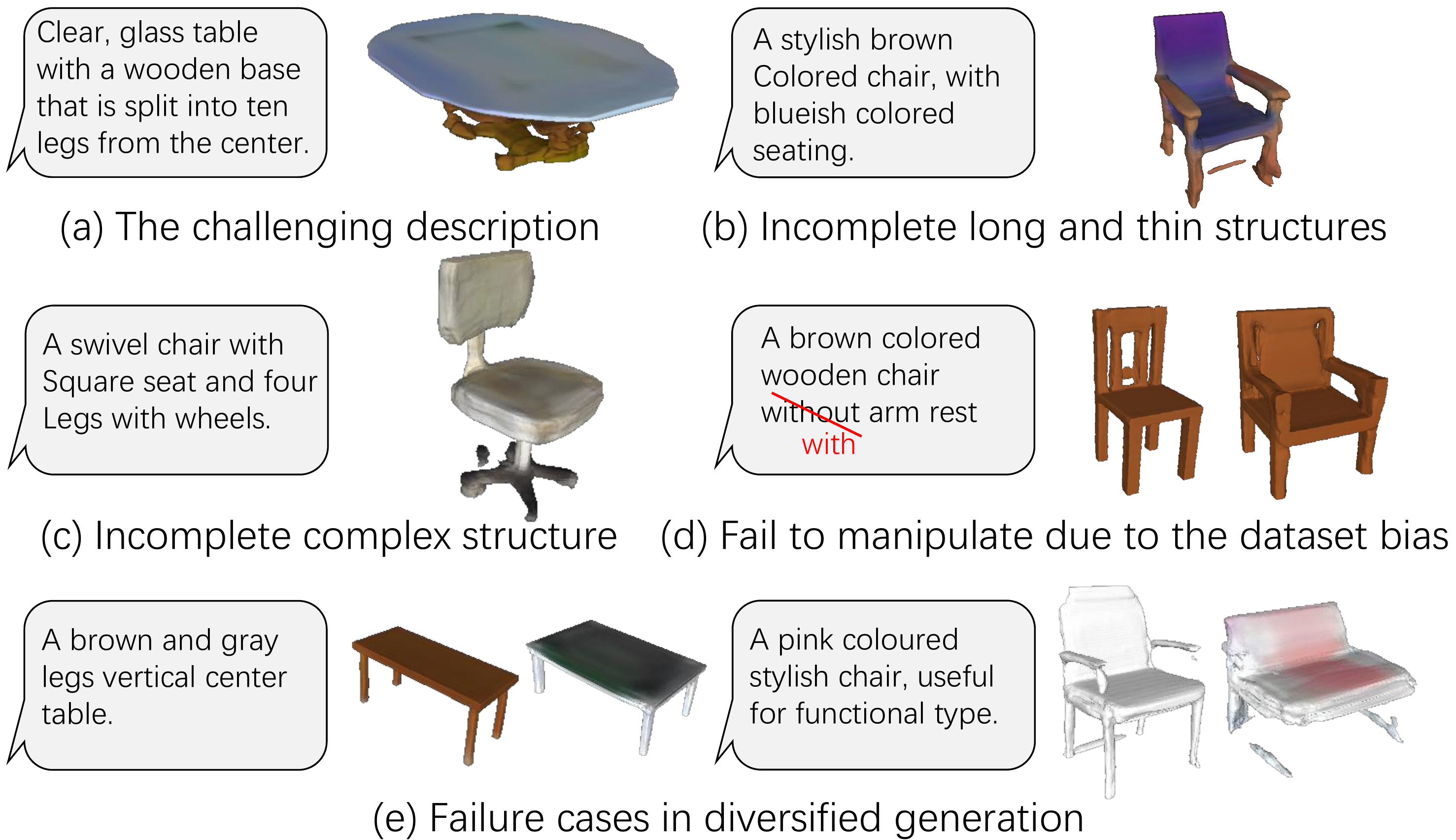}
\caption{Illustrating the limitations of our current approach.}
\label{fig:limitation}
\vspace*{-2mm}
\end{figure}
Our current approach still has some limitations.
First, text-guided shape generation is a very challenging task as discussed in Section 2 of the main paper. For example, some attributes, such as ``ten legs'' in Figure~\ref{fig:limitation} (a), are extremely challenging to generate. Our framework fails to generate a shape that faithfully follows such description.
Second, shapes with long, thin, and fine structures may get distorted or become noisy, as shown in Figure~\ref{fig:limitation} (b, c). To address this issue, we plan to explore more recent 3D implicit representations~\cite{li2020d,genova2020local,chen2021multiresolution}.
%
Also, the manipulation performance is limited by the inherent bias of the dataset. If we add an armrest to the chair in Figure~\ref{fig:limitation} (d), the manipulated chair will have become wider than the original one. Such a result is partially due to the dataset bias, since armed chairs are typically wider (like sofa) than those without armrests in the dataset. To resolve it, the manipulation process requires a topological understanding of the shapes. 
In addition, our metrics have some limitations. On the one hand, IoU may not be a good metric for text-to-shape generation task, because a slight difference in height/position between the generated shapes and GT shape can cause a low IoU; particularly, this is beyond the representative ability of a small piece of text. On the other hand, PS and FPD cannot fully reflect the generative quality, because these two metrics are based on the ScanObjectNN dataset~\cite{uy2019revisiting}, which has a large domain gap from our training dataset ShapeNet. In other words, better PS and FPD simply indicate that the generated shapes are more similar to the ScanObjectNN shapes, not necessarily meaning better quality. 
Also, there is a trade-off between the diversity and fidelity in our diversified generation. When stronger noise added to encourage the diversity, the quality of some generations cannot be ensured, and some are inconsistent with the text description as shown in Figure~\ref{fig:limitation} (e). It gets more serious when the text is long and contains descriptions on shape details. 
Last, our approach needs paired text-shape data for training, so we temporarily only explore shapes of table and chair, since the largest existing dataset~\cite{chen2018text2shape} only provides samples of these two categories. In the future, we will plan to explore zero-shot text-guided shape generation to extend the applicability of this work.
{\small
\bibliographystyle{ieee_fullname}
\bibliography{egbib}

\begin{thebibliography}{10}\itemsep=-1pt

\bibitem{abdal2019image2stylegan}
Rameen Abdal, Yipeng Qin, and Peter Wonka.
\newblock {Image2StyleGAN}: How to embed images into the {StyleGAN} latent
  space?
\newblock In {\em ICCV}, 2019.

\bibitem{achlioptas2018learning2}
Panos Achlioptas, Olga Diamanti, Ioannis Mitliagkas, and Leonidas~J. Guibas.
\newblock Learning representations and generative models for {3D} point clouds.
\newblock In {\em ICML}, 2018.

\bibitem{achlioptas2018learning}
Panos Achlioptas, Judy~E. Fan, Robert~X.D. Hawkins, Noah~D. Goodman, and
  Leonidas~J. Guibas.
\newblock Learning to refer to {3D} objects with natural language.
\newblock 2018.

\bibitem{arjovsky2017towards}
Martin Arjovsky and L{\'e}on Bottou.
\newblock Towards principled methods for training generative adversarial
  networks.
\newblock {\em ICLR}, 2017.

\bibitem{arora2021shape}
Himanshu Arora, Saurabh Mishra, Shichong Peng, Ke Li, and Ali Mahdavi-Amiri.
\newblock Shape completion via {IMLE}.
\newblock {\em arXiv preprint arXiv:2106.16237}, 2021.

\bibitem{ba2016layer}
Jimmy~Lei Ba, Jamie~Ryan Kiros, and Geoffrey~E. Hinton.
\newblock Layer normalization.
\newblock {\em arXiv preprint arXiv:1607.06450}, 2016.

\bibitem{brock2018large}
Andrew Brock, Jeff Donahue, and Karen Simonyan.
\newblock Large scale {GAN} training for high fidelity natural image synthesis.
\newblock {\em ICLR}, 2019.

\bibitem{cai2020learning}
Ruojin Cai, Guandao Yang, Hadar Averbuch-Elor, Zekun Hao, Serge Belongie, Noah
  Snavely, and Bharath Hariharan.
\newblock Learning gradient fields for shape generation.
\newblock In {\em ECCV}, 2020.

\bibitem{shapenet2015}
Angel~X. Chang, Thomas Funkhouser, Leonidas~J. Guibas, Pat Hanrahan, Qixing
  Huang, Zimo Li, Silvio Savarese, Manolis Savva, Shuran Song, Hao Su,
  Jianxiong Xiao, Li Yi, and Fisher Yu.
\newblock {ShapeNet: An Information-Rich {3D} Model Repository}.
\newblock Technical Report arXiv:1512.03012 [cs.GR], 2015.

\bibitem{chen2020scanrefer}
Dave~Zhenyu Chen, Angel~X. Chang, and Matthias Nie{\ss}ner.
\newblock {ScanRefer}: {3D} object localization in {RGB-D} scans using natural
  language.
\newblock In {\em ECCV}, 2020.

\bibitem{chen2018text2shape}
Kevin Chen, Christopher~B. Choy, Manolis Savva, Angel~X. Chang, Thomas
  Funkhouser, and Silvio Savarese.
\newblock {Text2Shape}: Generating shapes from natural language by learning
  joint embeddings.
\newblock In {\em ACCV}, 2018.

\bibitem{chen2021decor}
Zhiqin Chen, Vladimir~G. Kim, Matthew Fisher, Noam Aigerman, Hao Zhang, and
  Siddhartha Chaudhuri.
\newblock {DECOR-GAN}: {3D} shape detailization by conditional refinement.
\newblock In {\em CVPR}, 2021.

\bibitem{chen2020bsp}
Zhiqin Chen, Andrea Tagliasacchi, and Hao Zhang.
\newblock {BSP-Net}: Generating compact meshes via binary space partitioning.
\newblock In {\em CVPR}, 2020.

\bibitem{chen2019learning}
Zhiqin Chen and Hao Zhang.
\newblock Learning implicit fields for generative shape modeling.
\newblock In {\em CVPR}, 2019.

\bibitem{chen2021multiresolution}
Zhang Chen, Yinda Zhang, Kyle Genova, Sean Fanello, Sofien Bouaziz, Christian
  Hane, Ruofei Du, Cem Keskin, Thomas Funkhouser, and Danhang Tang.
\newblock Multiresolution deep implicit functions for 3d shape representation.
\newblock In {\em ICCV}, 2021.

\bibitem{chibane2020implicit}
Julian Chibane, Thiemo Alldieck, and Gerard Pons-Moll.
\newblock Implicit functions in feature space for {3D} shape reconstruction and
  completion.
\newblock In {\em CVPR}, 2020.

\bibitem{chibane2020neural}
Julian Chibane, Aymen Mir, and Gerard Pons-Moll.
\newblock Neural unsigned distance fields for implicit function learning.
\newblock In {\em NeurIPS}, 2020.

\bibitem{chibane2020implicit2}
Julian Chibane and Gerard Pons-Moll.
\newblock Implicit feature networks for texture completion from partial {3D}
  data.
\newblock In {\em ECCV}, 2020.

\bibitem{choy20163d}
Christopher~B. Choy, Danfei Xu, JunYoung Gwak, Kevin Chen, and Silvio Savarese.
\newblock {3D-R2N2}: A unified approach for single and multi-view {3D} object
  reconstruction.
\newblock In {\em ECCV}, 2016.

\bibitem{deng2021deformed}
Yu Deng, Jiaolong Yang, and Xin Tong.
\newblock Deformed implicit field: Modeling {3D} shapes with learned dense
  correspondence.
\newblock In {\em CVPR}, 2021.

\bibitem{devlin2018bert}
Jacob Devlin, Ming-Wei Chang, Kenton Lee, and Kristina Toutanova.
\newblock {BERT}: Pre-training of deep bidirectional transformers for language
  understanding.
\newblock {\em NAACL-HLT}, 2018.

\bibitem{feng2019meshnet}
Yutong Feng, Yifan Feng, Haoxuan You, Xibin Zhao, and Yue Gao.
\newblock {MeshNet}: Mesh neural network for {3D} shape representation.
\newblock In {\em AAAI}, 2019.

\bibitem{gao2019sdm}
Lin Gao, Jie Yang, Tong Wu, Yu-Jie Yuan, Hongbo Fu, Yu-Kun Lai, and Hao Zhang.
\newblock {SDM-NET}: Deep generative network for structured deformable mesh.
\newblock {\em ACM TOG (SIGGRAPH Asia)}, 2019.

\bibitem{genova2020local}
Kyle Genova, Forrester Cole, Avneesh Sud, Aaron Sarna, and Thomas Funkhouser.
\newblock Local deep implicit functions for {3D} shape.
\newblock In {\em CVPR}, 2020.

\bibitem{girdhar2016learning}
Rohit Girdhar, David~F. Fouhey, Mikel Rodriguez, and Abhinav Gupta.
\newblock Learning a predictable and generative vector representation for
  objects.
\newblock In {\em ECCV}, 2016.

\bibitem{han2020shapecaptioner}
Zhizhong Han, Chao Chen, Yu-Shen Liu, and Matthias Zwicker.
\newblock {ShapeCaptioner}: Generative caption network for {3D} shapes by
  learning a mapping from parts detected in multiple views to sentences.
\newblock In {\em ACM MM}, 2020.

\bibitem{han2019y2seq2seq}
Zhizhong Han, Mingyang Shang, Xiyang Wang, Yu-Shen Liu, and Matthias Zwicker.
\newblock Y2seq2seq: Cross-modal representation learning for {3D} shape and
  text by joint reconstruction and prediction of view and word sequences.
\newblock In {\em AAAI}, 2019.

\bibitem{hao2020dualsdf}
Zekun Hao, Hadar Averbuch-Elor, Noah Snavely, and Serge Belongie.
\newblock {DualSDF}: Semantic shape manipulation using a two-level
  representation.
\newblock In {\em CVPR}, 2020.

\bibitem{heusel2017gans}
Martin Heusel, Hubert Ramsauer, Thomas Unterthiner, Bernhard Nessler, and Sepp
  Hochreiter.
\newblock {GANs} trained by a two time-scale update rule converge to a local
  {N}ash equilibrium.
\newblock {\em NIPS}, 2017.

\bibitem{huang2021di}
Jiahui Huang, Shi-Sheng Huang, Haoxuan Song, and Shi-Min Hu.
\newblock {DI-Fusion}: Online implicit {3D} reconstruction with deep priors.
\newblock In {\em CVPR}, 2021.

\bibitem{hui2020progressive}
Le Hui, Rui Xu, Jin Xie, Jianjun Qian, and Jian Yang.
\newblock Progressive point cloud deconvolution generation network.
\newblock In {\em ECCV}, 2020.

\bibitem{huq2020static}
Faria Huq, Nafees Ahmed, and Anindya Iqbal.
\newblock Static and animated {3D} scene generation from free-form text
  descriptions.
\newblock {\em arXiv preprint arXiv:2010.01549}, 2020.

\bibitem{ibing20213d}
Moritz Ibing, Isaak Lim, and Leif Kobbelt.
\newblock {3D} shape generation with grid-based implicit functions.
\newblock In {\em CVPR}, 2021.

\bibitem{jahan2021semantics}
Tansin Jahan, Yanran Guan, and Oliver van Kaick.
\newblock Semantics-guided latent space exploration for shape generation.
\newblock In {\em COMPUT GRAPH FORUM}, 2021.

\bibitem{jiang2020shapeflow}
Chiyu Jiang, Jingwei Huang, Andrea Tagliasacchi, Leonidas~J. Guibas, et~al.
\newblock Shape{F}low: Learnable deformations among {3D} shapes.
\newblock {\em NeurIPS}, 2020.

\bibitem{jiang2020local}
Chiyu Jiang, Avneesh Sud, Ameesh Makadia, Jingwei Huang, Matthias Nie{\ss}ner,
  Thomas Funkhouser, et~al.
\newblock Local implicit grid representations for {3D} scenes.
\newblock In {\em CVPR}, 2020.

\bibitem{karras2017progressive}
Tero Karras, Timo Aila, Samuli Laine, and Jaakko Lehtinen.
\newblock Progressive growing of {GANs} for improved quality, stability, and
  variation.
\newblock {\em ICLR}, 2018.

\bibitem{karras2019style}
Tero Karras, Samuli Laine, and Timo Aila.
\newblock A style-based generator architecture for generative adversarial
  networks.
\newblock In {\em CVPR}, 2019.

\bibitem{kim2020softflow}
Hyeongju Kim, Hyeonseung Lee, Woo~Hyun Kang, Joun~Yeop Lee, and Nam~Soo Kim.
\newblock {SoftFlow}: Probabilistic framework for normalizing flow on
  manifolds.
\newblock {\em NeurIPS}, 2020.

\bibitem{klokov2020discrete}
Roman Klokov, Edmond Boyer, and Jakob Verbeek.
\newblock Discrete point flow networks for efficient point cloud generation.
\newblock In {\em ECCV}, 2020.

\bibitem{li2019controllable}
Bowen Li, Xiaojuan Qi, Thomas Lukasiewicz, and Philip H.~S. Torr.
\newblock Controllable text-to-image generation.
\newblock {\em NeurIPS}, 2019.

\bibitem{li2020manigan}
Bowen Li, Xiaojuan Qi, Thomas Lukasiewicz, and Philip H.~S. Torr.
\newblock {ManiGAN}: Text-guided image manipulation.
\newblock In {\em CVPR}, 2020.

\bibitem{li2020multimodal}
Ke Li, Shichong Peng, Tianhao Zhang, and Jitendra Malik.
\newblock Multimodal image synthesis with conditional implicit maximum
  likelihood estimation.
\newblock {\em IJCV}, 2020.

\bibitem{li2019diverse}
Ke Li, Tianhao Zhang, and Jitendra Malik.
\newblock Diverse image synthesis from semantic layouts via conditional {IMLE}.
\newblock In {\em ICCV}, 2019.

\bibitem{li2020d}
Manyi Li and Hao Zhang.
\newblock {D}$^2${IM-Net}: Learning detail disentangled implicit fields from
  single images.
\newblock {\em CVPR}, 2021.

\bibitem{li2021sp}
Ruihui Li, Xianzhi Li, Ka-Hei Hui, and Chi-Wing Fu.
\newblock {SP-GAN}: sphere-guided {3D} shape generation and manipulation.
\newblock {\em ACM TOG (SIGGRAPH)}, 2021.

\bibitem{lin2014microsoft}
Tsung-Yi Lin, Michael Maire, Serge Belongie, James Hays, Pietro Perona, Deva
  Ramanan, Piotr Doll{\'a}r, and C.~Lawrence Zitnick.
\newblock Microsoft {COCO}: Common objects in context.
\newblock In {\em ECCV}, 2014.

\bibitem{liu2021deep}
Shi-Lin Liu, Hao-Xiang Guo, Hao Pan, Peng-Shuai Wang, Xin Tong, and Yang Liu.
\newblock Deep implicit moving least-squares functions for {3D} reconstruction.
\newblock In {\em CVPR}, 2021.

\bibitem{mescheder2019occupancy}
Lars Mescheder, Michael Oechsle, Michael Niemeyer, Sebastian Nowozin, and
  Andreas Geiger.
\newblock Occupancy networks: Learning {3D} reconstruction in function space.
\newblock In {\em CVPR}, 2019.

\bibitem{mo2019structurenet}
Kaichun Mo, Paul Guerrero, Li Yi, Hao Su, Peter Wonka, Niloy Mitra, and
  Leonidas~J. Guibas.
\newblock Structure{N}et.: Hierarchical graph networks for {3D} shape
  generation.
\newblock {\em ACM TOG (SIGGRAPH Asia)}, 2019.

\bibitem{niemeyer2020differentiable}
Michael Niemeyer, Lars Mescheder, Michael Oechsle, and Andreas Geiger.
\newblock Differentiable volumetric rendering: Learning implicit {3D}
  representations without {3D} supervision.
\newblock In {\em CVPR}, 2020.

\bibitem{nilsback2008automated}
Maria-Elena Nilsback and Andrew Zisserman.
\newblock Automated flower classification over a large number of classes.
\newblock In {\em 2008 Sixth Indian Conference on Computer Vision, Graphics \&
  Image Processing}, 2008.

\bibitem{oechsle2019texture}
Michael Oechsle, Lars Mescheder, Michael Niemeyer, Thilo Strauss, and Andreas
  Geiger.
\newblock Texture fields: Learning texture representations in function space.
\newblock In {\em ICCV}, 2019.

\bibitem{oechsle2020learning}
Michael Oechsle, Michael Niemeyer, Christian Reiser, Lars Mescheder, Thilo
  Strauss, and Andreas Geiger.
\newblock Learning implicit surface light fields.
\newblock In {\em 3DV}, 2020.

\bibitem{park2019deepsdf}
Jeong~Joon Park, Peter Florence, Julian Straub, Richard Newcombe, and Steven
  Lovegrove.
\newblock {DeepSDF}: Learning continuous signed distance functions for shape
  representation.
\newblock In {\em CVPR}, 2019.

\bibitem{paszke2019pytorch}
Adam Paszke, Sam Gross, Francisco Massa, Adam Lerer, James Bradbury, Gregory
  Chanan, Trevor Killeen, Zeming Lin, Natalia Gimelshein, Luca Antiga, et~al.
\newblock Py{T}orch: An imperative style, high-performance deep learning
  library.
\newblock {\em NeurIPS}, 2019.

\bibitem{patashnik2021styleclip}
Or Patashnik, Zongze Wu, Eli Shechtman, Daniel Cohen-Or, and Dani Lischinski.
\newblock {StyleCLIP}: Text-driven manipulation of {StyleGAN} imagery.
\newblock {\em ICCV}, 2021.

\bibitem{peng2020generating}
Shichong Peng and Ke Li.
\newblock Generating unobserved alternatives.
\newblock {\em ICLR}, 2021.

\bibitem{peng2020convolutional}
Songyou Peng, Michael Niemeyer, Lars Mescheder, Marc Pollefeys, and Andreas
  Geiger.
\newblock Convolutional occupancy networks.
\newblock {\em ECCV}, 2020.

\bibitem{poursaeed2020coupling}
Omid Poursaeed, Matthew Fisher, Noam Aigerman, and Vladimir~G. Kim.
\newblock Coupling explicit and implicit surface representations for generative
  {3D} modeling.
\newblock In {\em ECCV}, 2020.

\bibitem{qi2017pointnet}
Charles~R. Qi, Hao Su, Kaichun Mo, and Leonidas~J. Guibas.
\newblock {PointNet}: Deep learning on point sets for {3D} classification and
  segmentation.
\newblock In {\em CVPR}, 2017.

\bibitem{qiao2019mirrorgan}
Tingting Qiao, Jing Zhang, Duanqing Xu, and Dacheng Tao.
\newblock {MirrorGAN}: Learning text-to-image generation by redescription.
\newblock In {\em CVPR}, 2019.

\bibitem{reed2016generative}
Scott Reed, Zeynep Akata, Xinchen Yan, Lajanugen Logeswaran, Bernt Schiele, and
  Honglak Lee.
\newblock Generative adversarial text to image synthesis.
\newblock In {\em ICML}, 2016.

\bibitem{reed2016learning}
Scott~E. Reed, Zeynep Akata, Santosh Mohan, Samuel Tenka, Bernt Schiele, and
  Honglak Lee.
\newblock Learning what and where to draw.
\newblock {\em NIPS}, 2016.

\bibitem{rombach2020network}
Robin Rombach, Patrick Esser, and Bj{\"o}rn Ommer.
\newblock Network-to-network translation with conditional invertible neural
  networks.
\newblock {\em NeurIPS}, 2020.

\bibitem{salimans2016improved}
Tim Salimans, Ian Goodfellow, Wojciech Zaremba, Vicki Cheung, Alec Radford, and
  Xi Chen.
\newblock Improved techniques for training {GANs}.
\newblock {\em NIPS}, 2016.

\bibitem{shu20193d}
Dong~Wook Shu, Sung~Woo Park, and Junseok Kwon.
\newblock {3D} point cloud generative adversarial network based on tree
  structured graph convolutions.
\newblock In {\em ICCV}, 2019.

\bibitem{souza2020efficient}
Douglas~M. Souza, J{\^o}natas Wehrmann, and Duncan~D. Ruiz.
\newblock Efficient neural architecture for text-to-image synthesis.
\newblock In {\em IJCNN}, 2020.

\bibitem{stap2020conditional}
David Stap, Maurits Bleeker, Sarah Ibrahimi, and Maartje ter Hoeve.
\newblock Conditional image generation and manipulation for user-specified
  content.
\newblock {\em CVPRW}, 2020.

\bibitem{sun2020pointgrow}
Yongbin Sun, Yue Wang, Ziwei Liu, Joshua Siegel, and Sanjay Sarma.
\newblock Point{G}row: Autoregressively learned point cloud generation with
  self-attention.
\newblock In {\em WACV}, 2020.

\bibitem{szegedy2016rethinking}
Christian Szegedy, Vincent Vanhoucke, Sergey Ioffe, Jon Shlens, and Zbigniew
  Wojna.
\newblock Rethinking the inception architecture for computer vision.
\newblock In {\em CVPR}, 2016.

\bibitem{tang2021part2word}
Chuan Tang, Xi Yang, Bojian Wu, Zhizhong Han, and Yi Chang.
\newblock {Part2Word}: Learning joint embedding of point clouds and text by
  matching parts to words.
\newblock {\em arXiv preprint arXiv:2107.01872}, 2021.

\bibitem{tretschk2020patchnets}
Edgar Tretschk, Ayush Tewari, Vladislav Golyanik, Michael Zollh{\"o}fer,
  Carsten Stoll, and Christian Theobalt.
\newblock {PatchNets}: Patch-based generalizable deep implicit {3D} shape
  representations.
\newblock In {\em ECCV}, 2020.

\bibitem{uy2019revisiting}
Mikaela~Angelina Uy, Quang-Hieu Pham, Binh-Son Hua, Thanh Nguyen, and Sai-Kit
  Yeung.
\newblock Revisiting point cloud classification: A new benchmark dataset and
  classification model on real-world data.
\newblock In {\em ICCV}, 2019.

\bibitem{wah2011caltech}
C. Wah, S. Branson, P. Welinder, P. Perona, and S. Belongie.
\newblock {The Caltech-UCSD Birds-200-2011 Dataset}.
\newblock Technical Report CNS-TR-2011-001, California Institute of Technology,
  2011.

\bibitem{wang2021cycle}
Hao Wang, Guosheng Lin, Steven Hoi, and Chunyan Miao.
\newblock Cycle-consistent inverse {GAN} for text-to-image synthesis.
\newblock {\em ACM MM}, 2021.

\bibitem{wang2020text}
Zixu Wang, Zhe Quan, Zhi-Jie Wang, Xinjian Hu, and Yangyang Chen.
\newblock Text to image synthesis with bidirectional generative adversarial
  network.
\newblock In {\em ICME}, 2020.

\bibitem{wu2020pq}
Rundi Wu, Yixin Zhuang, Kai Xu, Hao Zhang, and Baoquan Chen.
\newblock {PQ-NET}: A generative part seq2seq network for {3D} shapes.
\newblock In {\em CVPR}, 2020.

\bibitem{xia2021tedigan}
Weihao Xia, Yujiu Yang, Jing-Hao Xue, and Baoyuan Wu.
\newblock {TediGAN}: Text-guided diverse face image generation and
  manipulation.
\newblock In {\em CVPR}, 2021.

\bibitem{xu2019disn}
Qiangeng Xu, Weiyue Wang, Duygu Ceylan, Radomir Mech, and Ulrich Neumann.
\newblock {DISN}: Deep implicit surface network for high-quality single-view
  {3D} reconstruction.
\newblock {\em NeurIPS}, 2019.

\bibitem{xu2018attngan}
Tao Xu, Pengchuan Zhang, Qiuyuan Huang, Han Zhang, Zhe Gan, Xiaolei Huang, and
  Xiaodong He.
\newblock Attn{GAN}: Fine-grained text to image generation with attentional
  generative adversarial networks.
\newblock In {\em CVPR}, 2018.

\bibitem{Jie20DsgNet}
Jie Yang, Kaichun Mo, Yu-Kun Lai, Leonidas~J. Guibas, and Lin Gao.
\newblock {DSG-Net}: Learning disentangled structure and geometry for {3D}
  shape generation.
\newblock {\em ACM TOG (SIGGRAPH Asia)}, 2021.

\bibitem{yariv2020multiview}
Lior Yariv, Yoni Kasten, Dror Moran, Meirav Galun, Matan Atzmon, Ronen Basri,
  and Yaron Lipman.
\newblock Multiview neural surface reconstruction by disentangling geometry and
  appearance.
\newblock {\em NeurIPS}, 2020.

\bibitem{yifan2021iso}
Wang Yifan, Shihao Wu, Cengiz Oztireli, and Olga Sorkine-Hornung.
\newblock {Iso-Points}: Optimizing neural implicit surfaces with hybrid
  representations.
\newblock In {\em CVPR}, 2021.

\bibitem{yuan2019bridge}
Mingkuan Yuan and Yuxin Peng.
\newblock Bridge-{GAN}: Interpretable representation learning for text-to-image
  synthesis.
\newblock {\em IEEE TCSVT}, 2019.

\bibitem{zhang2017stackgan}
Han Zhang, Tao Xu, Hongsheng Li, Shaoting Zhang, Xiaogang Wang, Xiaolei Huang,
  and Dimitris~N. Metaxas.
\newblock {StackGAN}: Text to photo-realistic image synthesis with stacked
  generative adversarial networks.
\newblock In {\em ICCV}, 2017.

\bibitem{zhang2018stackgan++}
Han Zhang, Tao Xu, Hongsheng Li, Shaoting Zhang, Xiaogang Wang, Xiaolei Huang,
  and Dimitris~N. Metaxas.
\newblock {StackGAN}++: Realistic image synthesis with stacked generative
  adversarial networks.
\newblock {\em IEEE TPAMI}, 2018.

\bibitem{zheng2021deep}
Zerong Zheng, Tao Yu, Qionghai Dai, and Yebin Liu.
\newblock Deep implicit templates for {3D} shape representation.
\newblock In {\em CVPR}, 2021.

\end{thebibliography}
}

\end{document}